\UseRawInputEncoding 
\documentclass[review,authoryear]{elsarticle}
\usepackage{lineno,hyperref}\modulolinenumbers[5]

\usepackage{appendix}
\usepackage{amsmath}
\usepackage{multicol}
\usepackage{subfigure}
\usepackage{blindtext}
\usepackage{tabularx}
\usepackage{algorithm, algpseudocode}
\usepackage{enumitem}
\usepackage{rotating}
\usepackage{pdflscape}
\usepackage{adjustbox}
\usepackage{amsmath}
\usepackage{amssymb}
\usepackage{mathtools}
\usepackage{pdfpages}
\usepackage{xcolor}

\DeclarePairedDelimiter\floor{\lfloor}{\rfloor}

\journal{Engineering Applications of Artificial Intelligence}

\biboptions{round,sort,authoryear}

\begin{document}

\begin{frontmatter}

\title{Enabling Integration and Interaction for Decentralized Artificial Intelligence in Airline Disruption Management \tnoteref{mytitlenote}}
\tnotetext[mytitlenote]{This article represents a chapter from the corresponding author's completed doctoral dissertation.}

\author[1]{Kolawole Ogunsina\corref{cor1}%
\fnref{fn1}}
\ead{kolawole08@gmail.com}

\author[2]{Daniel DeLaurentis\fnref{fn2}}
\ead{ddelaure@purdue.edu}

\cortext[cor1]{Corresponding Author}
\fntext[fn1]{School of Aeronautics and Astronautics, Purdue University, West Lafayette, IN 47907, United States.}
\fntext[fn2]{School of Aeronautics and Astronautics, Purdue University, West Lafayette, IN 47907, United States.}





\begin{abstract}
Airline disruption management traditionally seeks to address three problem dimensions: aircraft scheduling, crew scheduling, and passenger scheduling, in that order. However, current efforts have, at most, only addressed the first two problem dimensions concurrently and do not account for the propagative effects that uncertain scheduling outcomes in one dimension can have on another dimension. In addition, existing approaches for airline disruption management include human specialists who decide on necessary corrective actions for airline schedule disruptions on the day of operation. However, human specialists are limited in their ability to process copious amounts of information imperative for making robust decisions that simultaneously address all problem dimensions during disruption management. Therefore, there is a need to augment the decision-making capabilities of a human specialist with quantitative and qualitative tools that can rationalize complex interactions amongst all dimensions in airline disruption management, and provide objective insights to the specialists in the airline operations control center. To that effect, we provide a discussion and demonstration of an agnostic and systematic paradigm for enabling expeditious simultaneously-integrated recovery of all problem dimensions during airline disruption management, through an intelligent multi-agent system that employs principles from artificial intelligence and distributed ledger technology. Results indicate that our paradigm for simultaneously-integrated recovery executes in polynomial time and is effective when all the flights in the airline route network are disrupted. 
\end{abstract}


\end{frontmatter}


\section{Introduction}

\begin{figure}[ht!]
	\centering
	\includegraphics[width=0.99\textwidth]{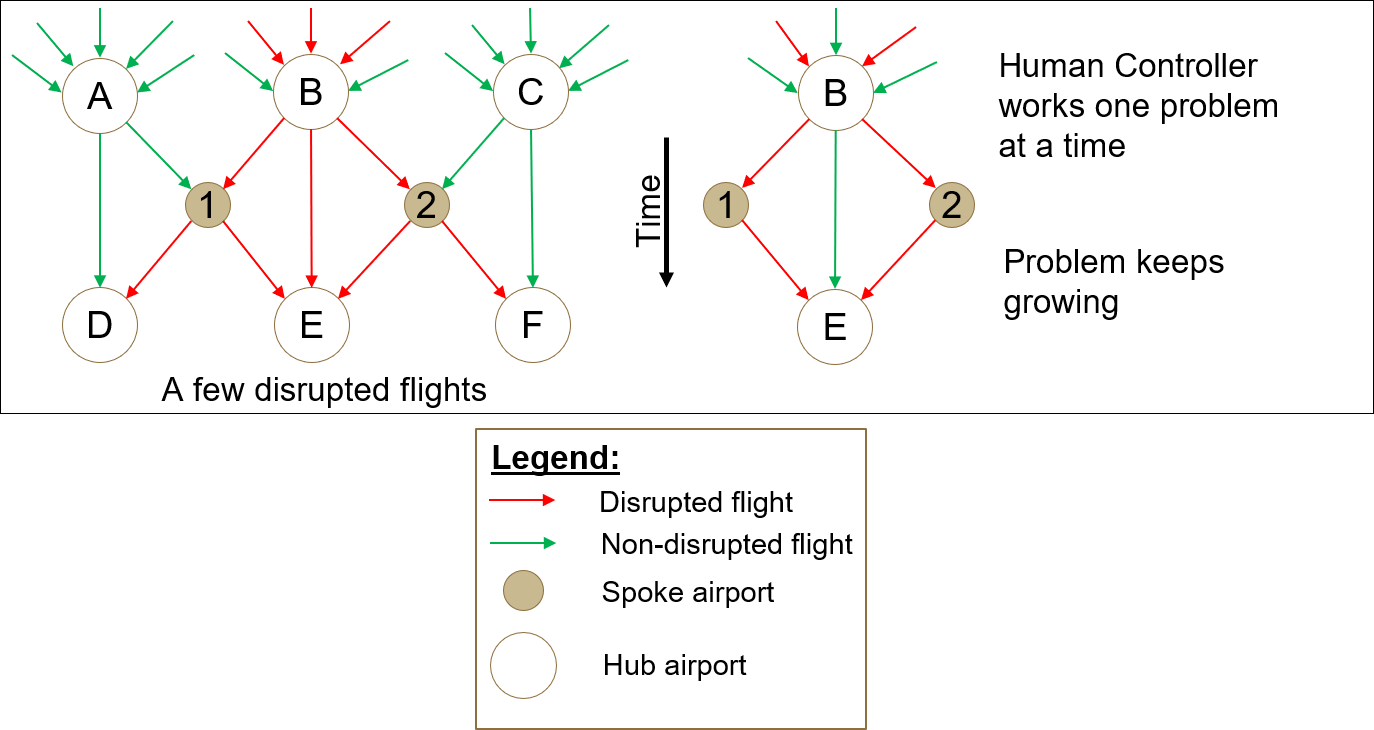}
	\caption{Human involvement in disruption management and operations recovery at the AOCC}
	\label{fig:IROPS_Impetus}
\end{figure}

Humans assume a primary role in the development, evolution, and assessment of existing paradigms and systems adopted for making decisions during airline operations recovery and disruption management \citep{Deloitte2017}. Current paradigms for operations recovery at many airlines involve database query systems, which allow human operators (or specialists) in the airline operations control center (AOCC) to perform inquires on databases in order to effectively assess solutions proffered by decision support systems for different problem dimensions (i.e. aircraft scheduling, crew scheduling, and passenger scheduling) during irregular operations and disruption management. Together, database query systems and decision support systems provide a credible platform for addressing irregular operations, as shown in Fig.~\ref{fig:IROPS_Impetus}, during schedule execution on day of operation.

\subsection{The Problem}
The efficiency of a platform defined by Fig.~\ref{fig:IROPS_Impetus} is bottlenecked by human operators for two reasons. First, humans find it difficult to simultaneously utilize large volumes of data to make the best decision that spans multiple problem  dimensions for disruption management \citep{Castro2006, Mosier2018}. In other words, they can not promptly employ all available and necessary data to select and apply appropriate information that is most effective for managing irregular operations. Second, plausibly inferior quality (i.e. goodness) of information (retrieved by the human operator from a database query system) that is subsequently passed on to a decision support system can compromise the efficacy of solutions provided by the decision support system, thus rendering disruption resolutions sub-optimal. To complicate decision-making matters, existing paradigms for airline disruption management do not readily allow human experts to see the impact of these disruption resolutions (partly informed by database query systems) until the recovery plan is already in motion and its consequence can not be revoked \citep{Deloitte2017}. As such, delineating skill from luck while managing disruptions for optimal airline recovery remains an open and challenging problem. To this effect, current industry techniques for airline disruption management are unable to attain and maintain expeditious, effective, and concurrent recovery of all problem dimensions in a disrupted airline route network. 

Furthermore, many decision support systems adopted by current integrated recovery paradigms for airline disruption management are imbued in a monolithic system design doctrine, wherein specifications are created first before a system that meets the specifications is constructed. However, this design approach has not succeeded in eradicating irregular airline operations because specifications continue to evolve as new capabilities are added to an existing system \citep{Amadeus2016}. Although current approaches for airline disruption management are capable of providing decision support, albeit restricted by a strict sequential resolution process for problem dimensions, the rapid evolutionary manner in which different situations for irregular operations occur typically renders solutions from decision support systems ineffectual within moments after they are generated. Moreover, adding more proficiency, in form of computerized information systems, to existing decision support systems in the AOCC to address fleeting solution validity significantly increases the complexity of the monolithic and centralized design approach for airline disruption management. To this end, there is a need to substantially augment the decision-making capacity of human specialists in the AOCC with modular and decentralized decision support platforms that allow real-time generation and tracking of disruption resolutions with minimal complexity during irregular operations and schedule recovery.   

Recent advancements in artificial intelligence (AI) and distributed ledger technology (DLT) \citep{Castro2002,Swan2015,Heaton2015,Choi2018,Maxmen2018,Baird2018a} have provided an avenue to develop decision support systems that can allow human specialists in the AOCC to readily assess and validate the effectiveness of their decisions while concurrently recovering the airline network during irregular operations. To that effect, this paper provides a compendious discussion of the mechanisms that enable the integration of constituent AI models developed from our previous work \citep{Ogunsina2021a, Ogunsina2021b}, which define the intelligent agent for each functional role (or domain) in the AOCC, and the interaction of multiple intelligent agents for simultaneously-integrated recovery during airline disruption management.

\subsection{Paper Organization}
We begin our discussion in Section \ref{back_motiv} by describing existing practices for airline disruption management and the motivation for our research contributions. Next, Section \ref{AI_DLT} elucidates different tenets of AI and DLT, which provide a conducive medium for achieving scalable simultaneously-integrated recovery of airline schedule during disruption management. We describe the consensus platform in Section \ref{case_study} for a special type of distributed ledger called Hashgraph, and how routines that define separate AI models serve as elements for a proof-of-stake system to achieve simultaneously-integrated recovery. Section \ref{results} discusses a demonstration of the AI-DLT synthesis for simultaneously-integrated recovery of sets of disrupted flight schedules across multiple functional roles in the AOCC at a major U.S. airline. We conclude with a summary of pertinent findings and areas for further research in Section \ref{conclusion}.

\section{Background and Motivation}\label{back_motiv}
\subsection{Current Practice for Airline Disruption Management}
Existing resolution approaches for addressing the airline operations recovery problem, during disruption management, have generally focused on using traditional operations research (i.e. optimization and heuristics) methods that resolve one or more (but not all) of the primary problem dimensions. These schedule recovery approaches try to achieve one or more of a set of possible objectives, such as: minimizing the cost of reserve crews and spare aircraft used; minimizing passenger recovery costs; minimizing loss of passenger goodwill; and minimizing the amount of time until it is possible to resume the original schedule \citep{Lan2006}. Regardless of the objective, the airline operations recovery problem must be solved within minutes, and this time limitation makes it unrealistic to solve large monolithic optimization models. As such, to meet these objectives, most airline recovery processes are sequential in nature \citep{Barnhart2009}. 
\begin{figure}[b]
	\centering
	\includegraphics[width=0.99\textwidth]{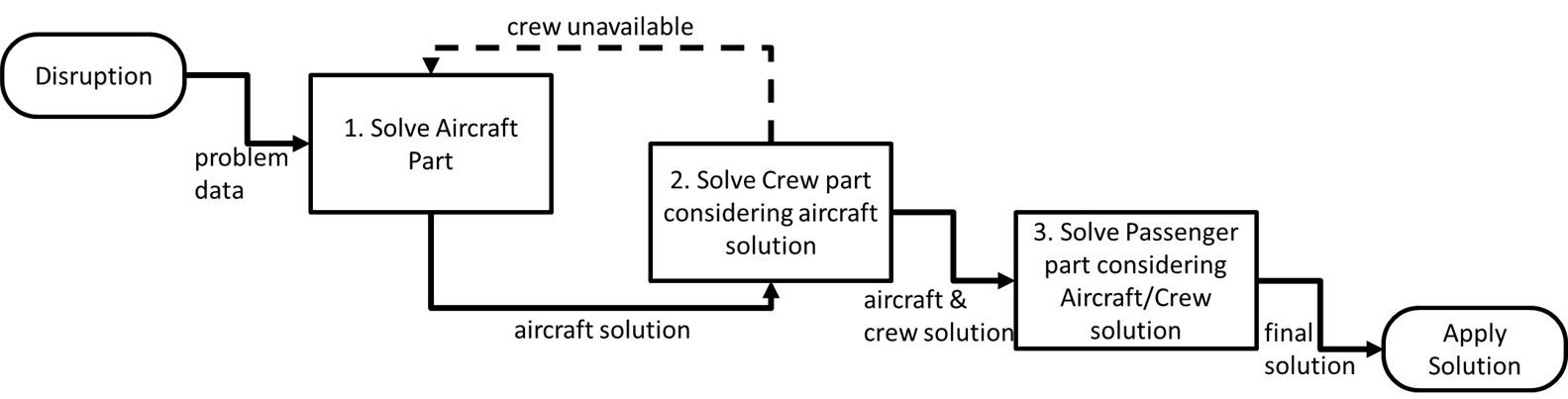}
	\caption{Current (sequential) practice for airline disruption management \citep{Castro2014}}
	\label{fig:CurrentFramework}
\end{figure} 

Fig.~\ref{fig:CurrentFramework} shows the current practice for airline disruption management. When a disruption in scheduled operations occurs, the AOCC typically reacts by resolving the problem in a sequential manner where issues related to the aircraft fleet, crew members, and passengers are addressed respectively, in that order, by their corresponding human specialists. These specialists, stationed in the AOCC, proactively monitor and mitigate problems and disruptions related to aircraft, crew members, and passengers in the airline network. The resolutions implemented by human specialists in each phase of the airline recovery process influence the resolutions applied in subsequent phases. Hence, the global objectives of the AOCC are achieved such that the overall airline recovery is restricted to a specific order during resolution (i.e. aircraft-crew-passenger). \cite{Petersen2012} were the first to present computational results on a fully integrated airline operations recovery problem, based upon the recovery practice shown in Fig.~\ref{fig:CurrentFramework}. The authors use a backtracking optimization approach, via a Benders decomposition scheme, to develop and solve a schedule recovery model. The schedule recovery model is the master problem that relates several variables from different sub-problems, which represent the problem dimensions in airline disruption management. However, a sub-problem resolution order is inherently imposed by the optimization algorithm, thereby making the aircraft and crew problem dimensions consistently more important than the passenger problem dimension. This approach was tested using data from a major US airline with a dense network, and showed to be effective when no more than 65\% of the flights in the airline route network are subject to disruption.

Furthermore, most decision support systems used to solve each problem dimension in the current recovery practice, shown in Fig.~\ref{fig:CurrentFramework}, are often deterministic and require the assessment of a human specialist before the resolutions generated by these support systems, at one or more phases of the recovery, are implemented as corrective action \citep{Clausen2010}. In addition to aleatoric uncertainty stemming from the random occurrence of disruptive events (like bad weather) on the day of operation, the current practice introduces epistemic uncertainty at each phase of the recovery when human specialists, with different levels of experience, are required to make decisions that will affect the solution generated in the subsequent phase.

According to the Amadeus IT group (a major IT group for the global travel industry), one of the main drivers that has significantly contributed to the lack of progress in developing a full solution to airline disruption management is limited bandwidth of human specialists \citep{Sousa2016}. Several key decisions at each phase of the recovery practice shown in Fig.~\ref{fig:CurrentFramework}, such as corrective actions implemented for a certain disruption type, are made by human specialists in the AOCC. Although human specialists are flexible in decision-making, they are not capable of parsing the large amounts of data necessary for simultaneously making robust real-time decisions for all the problem dimensions in airline operations recovery. In major airlines, adding more personnel to the AOCC does not effectively increase human bandwidth because of the significantly large size of the airline route network \citep{Sousa2016}. As such, there has been an emergence of a new class of resolution paradigms for airline disruption management, called simultaneously-integrated recovery \citep{Castro2014}, which aims to recover all problem dimensions simultaneously. Thus, the primary objective of simultaneously-integrated recovery is to remove the inherent constraint on the design and solution space for airline disruption management, which is imposed in the current recovery practice by solving the aircraft, crew, then passenger problem dimensions in that order. 

\subsection{Simultaneously-integrated Recovery Paradigm for Airline Disruption Management}
The simultaneously-integrated recovery paradigm is a design philosophy that enables the modeling of all support functions in the AOCC as a system of intelligent agents, such that their interactions are concurrently driven by local and global objectives. As such, this approach enables the realization of an acknowledged, distributed and scalable AOCC that can significantly reduce the problems encountered by existing AOCC organizations during disruption management. The acknowledged attribute of a new AOCC that adopts this design paradigm ensures that while different functional roles (or intelligent agents) are responsible for their respective local objectives, their actions are subordinate to improving the global objective of the AOCC \citep{Maier1998}. The distributed attribute is of three forms namely functional distribution, spatial distribution, and physical distribution \citep{Castro2014}. Functional distribution allows the existing roles and functions in the AOCC to be distributed and managed using intelligent software agents. Spatial distribution allows the data and information utilized by different roles or intelligent agents to be distributed. For instance, data in different databases and data in the same database but different partitions are spatially distributed. As such, this attribute can help alleviate system and data integration problems in the AOCC. Physical distribution ensures that the roles and functions designated to different intelligent agents can be distributed to different machines, such that the agents have access to more computational resources. Lastly, the scalability attribute allows the AOCC framework to grow as long as functional and spatial requirements are satisfied accordingly. Based upon the combined effect from these attributes, the AOCC is best represented as a multi-agent system \citep{Panait2005, Olfati-Saber2007}, such that support functions with frequently executed or repetitive tasks are performed by intelligent software agents. Some other benefits of adopting a multi-agent system framework for enabling simultaneously-integrated recovery in the AOCC include:

\begin{itemize}
    \item Ability to consider all problem dimensions at the same (or different) level of importance when generating resolutions (i.e. recovery plans).
    \item Increased autonomy and automation.
    \item Ability to measure local performance of constituent intelligent agents and global performance of the multi-agent system.
    \item Consideration of local preferences of each intelligent agent responsible for solving different problem dimensions.
    \item Ability to consider environment dynamics based upon the fact that existing information can change while resolutions are being generated.
    \item Ability to generate recovery plans in real or almost-real time.
\end{itemize}

There has been only one approach, by \cite{Castro2014}, that uses the simultaneously-integrated recovery paradigm till date. The authors use a multi-agent system design approach to characterize the AOCC at TAP Air Portugal, such that human roles in each problem dimension - with the most frequent tasks - are performed by intelligent agents. Their approach uses an adaptive protocol, called the Generic-Q-Negotiation (GQN) \citep{Watkins1992}, to ensure multi-attribute negotiation, with several rounds of feedback, between two types of agents (organizer and respondent) in order to achieve consensus. However, this disruption resolution approach is unable to handle evolving information dynamics (i.e. uncertainty in information evolution) during the disruption management process. Furthermore, the interaction amongst agents in the multi-agent system relies on a model-free learning environment, which does not employ a flight operations and scheduling model consistent with airline scheduling practices. As such, their approach requires significant time and trial-and-error experience, in form of many rounds of qualitative feedback between agents, to arrive at acceptable recovery plans during airline disruption management. Thus, there is need for a simultaneously-integrated recovery approach that requires little bandwidth to achieve consensus amongst intelligent agents. 

\subsection{Contributions}
We introduce an intelligent multi-agent system framework for airline disruption management that strictly applies historical data on airline scheduling and recovery operations to find optimum flight schedule recovery plans, which are attained through simultaneous interaction of functional roles (i.e. intelligent agents or domain managers) in the AOCC. To the best of our knowledge, our intelligent multi-agent system framework is the first architecture for airline disruption management that employs concurrent virtual voting amongst functional roles in the AOCC to achieve an automatic (or semi-automatic) system for simultaneously-integrated recovery. As such, our intelligent multi-agent system achieves consensus (i.e. Byzantine agreement) on any number of disruption resolutions without voting (feedback) on disruption resolutions, thereby ensuring zero bandwidth for communication of disruption resolutions amongst intelligent agents. To achieve the aforementioned objectives, this paper enhances prior research on simultaneously-integrated recovery for airline disruption management through the following contributions:
\begin{enumerate}
    \item We enable the fusion of AI and DLT for airline disruption management, by developing the intelligent multi-agent system architecture for multiple functional roles in the AOCC as a decentralized AI platform.
    \item We create a protocol to invoke the Hashgraph consensus algorithm, which employs the reliability of any disruption resolution provided by a human specialist as the stake (i.e. relative interest) of the associated functional role during its interaction with other participating functional roles in the AOCC.
    \item We establish the efficacy of a decentralized AI platform for simultaneously-integrated recovery in airline disruption management, by assessing scenarios of randomly disrupted flight schedules across multiple functional roles and problem dimensions in the AOCC. 
\end{enumerate}

\section{A Symbiotic Synthesis of AI and DLT}\label{AI_DLT}
One of the primary objectives of the fourth industrial revolution (i.e. Industry 4.0) is to transform traditional industrial practices by combining these practices with the latest smart technology \citep{Oztemel2020}. As such, current industry practices for irregular airline operations, which involve multiple air transportation stakeholders, can benefit amply from the use of machine to machine communication and internet of things (IoT) deployments to achieve increased automation, better communication, and self-monitoring during disruption management without significant need for human intervention. Furthermore, the unrelenting ubiquity of data that span many different areas of society has necessitated and enabled the creation of new interdisciplinary principles founded on mathematics, statistics, and probability theory, which enable machines (or computers) to have cognitive functions to learn, infer and adapt by leveraging data \citep{Jaynes2003,Piccarozzi2018}. To this effect, there is a strong mandate to explore decentralized interactions among intelligent machines that represent functional roles in the AOCC for better disruption management. 

\subsection{Artificial Intelligence}
Artificial intelligence or AI is the field that studies and applies interdisciplinary principles to achieve computationally intelligent agents that can directly assist humans with their day-to-day functions. A very prominent principle used in AI is machine learning \citep{Chui2018, Nguyen2018}, which relies on a centralized model for training wherein a group of servers run a particular algorithm against many training and validation examples, as described by \cite{Ogunsina2021}. As such, the few AI systems that exist (or in development) today for airline disruption management are generally specialized expert systems that utilize rolling and centralized data from a database query system to assist human specialists in the AOCC with making decisions during irregular operations \citep{Liu2019, Ye2020}.  Thus, current AI trends for airline disruption management aim to enable automated machine learning processes to manage data acquisition and knowledge updates for database query systems, thereby minimizing the manual labor required by human specialists for developing and evaluating decision support systems to mitigate irregular operations during schedule execution. Furthermore, the complex nature of the interaction between different actors (e.g. functional roles in the AOCC and air transportation stakeholders like air traffic controllers) does not augur well for robust airline disruption management, because a centralized AI system can not efficiently capture the proclivities amongst actors to discover and track emergent behavior in collective decision-making during schedule execution in a scalable manner \citep{Pomerol1997,Kiela2016, Li2018}. As such, a complementary platform is required to enhance the performance qualities of AI systems for improved disruption management during irregular operations. 

\subsection{Distributed Ledger Technology and Blockchain}
Distributed ledger technology or DLT \citep{Rauchs2018} represents an emerging technology that embodies two peculiar properties. First, it is distributed in nature such that the agreement about the state of a particular ledger being maintained is attained through recompensed consensus by a network of intelligent agents in lieu of relying on trust in a third-party intermediary that is extraneous to the network. Second, intelligent agents can deposit digital assets such as acts, timestamps, and states in the ledger, whose records are readily auditable, transparent, and incommutable based upon cryptographic and distributed basis that resist censorship and manipulation of assets \citep{maull2017}. To that effect, DLT enables high levels of anonymity and pseudonymity for intelligent agents during transactions (i.e. digital asset trades), such that records of transaction activities are observable at the meta-level and resistant to manipulation. However, the identification of specific intelligent agents performing trades on the ledger remains impossible during transactions, thereby rendering DLT immensely resilient to the intolerance of false information sharing for enhanced data security. The first noteworthy application of DLT was for the cryptocurrency named Bitcoin \citep{Nakamoto2008,Wright2019}. 

The blockchain, which is a rudimentary but effective form of DLT, was used to solve the problem of distributed consensus in a trustless network that provides a secure, controlled, and decentralized method of minting the Bitcoin digital currency. At its core, blockchain is a layer of a distributed peer-to-peer network running on the front end of the internet. In that regard, blockchain is characterized by: a data structure for tracking digital footprints in logical blocks of information, a shared ledger for recording actions (e.g. digital currency mining activities) by member computer systems or intelligent agents, and a decentralized consensus mechanism that solves a complex and random mathematical problem to validate member interactions based upon different routines that define engagement rules for participating members. DLT can be permissioned or permissionless depending on the manner in which access for information exchange is granted to current and prospective members (i.e. intelligent agents) of the DLT. Thus, a permissioned DLT only allows authorized intelligent agents to access the DLT application in private, consortium, or cloud-based settings, while a permissionless DLT like the Bitcoin blockchain is publicly accessible to any intelligent (mining) agent via the internet.  

\subsection{Decentralized AI for Airline Disruption Management}
\subsubsection{Background}
The existing scope of AI applications for airline disruption management primarily focuses on automated machine learning to improve the quality of information retrieved by human specialists from database query systems \citep{Liu2019}. While AI provides a means for improving the decision-making flexibility of human specialists in the AOCC during irregular operations, the centralized nature of existing monolithic systems for disruption managment does not enhance the efficiency of the decision-making process as a whole \citep{Petersen2012, Bouarfa2018}. Thus, agile decision-making for irregular operations during airline schedule execution requires a platform that inculcates decentralized intelligence during disruption management \citep{Janson2008}.

Furthermore, many existing integrated recovery paradigms for airline disruption management employ decision support mechanisms (i.e. explicit time-space optimization formulations) that provide disruption resolutions for separate problem dimensions without readily disclosing the unique rationale behind the solution generation process for a specific disruption event. As such, human specialists in the AOCC are often detached from the resolution creation process of the decision support system. In addition, majority of existing system design paradigms for airline operations recovery are \textit{Models and Algorithms} typically created by developers that operate separately from the AOCC organization. As such, \textit{Models and Algorithms} are not willingly accepted nor included in many highly customized tools and systems currently used by the AOCC for disruption management \citep{Ball2006, Castro2014, Bouarfa2018}.

A recent study of the Southwest Airlines Baker workgroup for disruption management, conducted by Deloitte Insights \citep{Deloitte2017}, revealed that having  superintendents of dispatch (i.e. human specialists in the AOCC) manage the creation of decision support systems that they themselves would use guaranteed that the disruption resolution sought by developers and users was shared and agreeable. This way, human specialists from different functional roles in the AOCC were more likely to accept and commit to a disruption resolution outcome proferred by a decision support system that is truly representative of their individual and collective decision-making expertise. As such, human specialists in the AOCC would prefer a decision support system that can expediently verify the efficacy and validity of their decision to implement a certain disruption resolution, in lieu of extant decision support systems that propose a disruption resolution to which the specialists have to decide whether or not to implement the proffered resolution. To that effect, a decentralized AI platform lends a suitable medium for developing next generation decision support systems for airline disruption management.  

\begin{figure}[b!]
	\centering
	\includegraphics[width=0.99\textwidth]{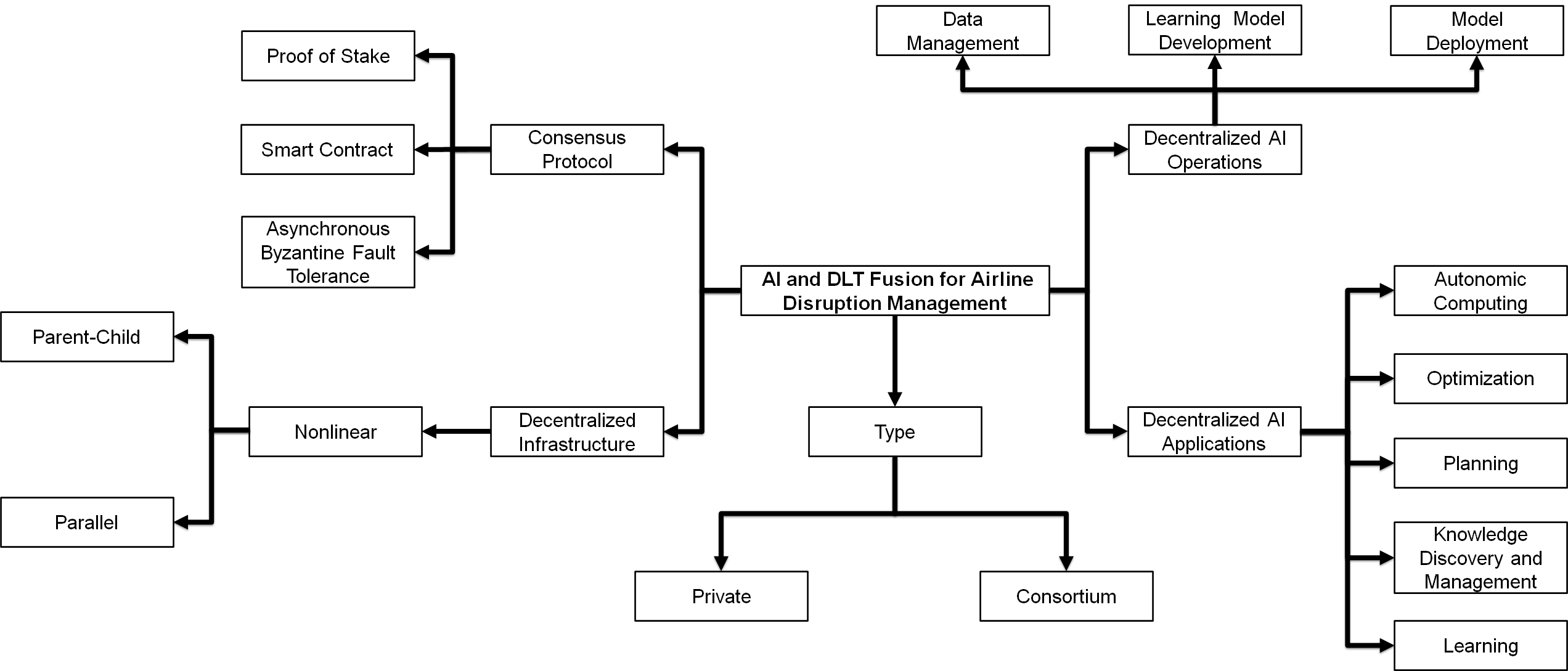}
	\caption{Taxonomy of a decentralized AI platform for airline disruption management}
	\label{fig:AI-DLT_tax}
\end{figure}

\subsubsection{Taxonomy}
The synthesis of AI and DLT (i.e. decentralized AI) can be categorized into two separate but related decentralized platforms, based upon certain integration properties and benefits, namely: \textit{Blockchain for AI} and \textit{AI for Blockchain} \citep{Dinh2018, Dai2019}. \textit{Blockchain for AI} represents a disruptive integration of AI and DLT that aims to help AI systems to attain the following enhancements: i) a secure data sharing environment for intelligent agents, ii) decentralized computing for intelligent agents, iii) explainable rationale for the actions of intelligent agents, iv) the coordination of distrusting intelligent agents. In complement, \textit{AI for Blockchain} represents an integration of AI and DLT that seeks to use AI to improve the automation, decision-making, and optimization of DLT (such as blockchain) for enhanced performance and governance. In that regard, \textit{AI for Blockchain} aims to achieve the following enhancements for DLT: i) secure and scalable distributed ledgers, ii) readily customizable distributed ledger systems that preserve the privacy of intelligent agents, iii) automated refereeing and governance of participating intelligent agents for distributed computing. A hybrid of these platforms represents a unique platform that enables multiple separate enhancements applicable to \textit{Blockchain for AI} and \textit{AI for Blockchain}. Fig.~\ref{fig:AI-DLT_tax} shows a tree diagram that represents five separate aspects of taxonomy in decentralized AI for airline disruption management based upon type, application, operation, consensus, and infrastructure \citep{Salah2019}.

\begin{itemize}
    \item \textbf{Type}: The type of decentralized AI platform for airline disruption management is defined by the importance and hierarchy (i.e. caliber) of stakeholders involved in irregular operations. As such, there are two appropriate types of decentralized AI namely: private and consortium AI systems respectively. A private decentralized AI for airline disruption management represents a \textit{Blockchain for AI} platform that models the intrinsic behavioral properties of individual high-level stakeholders (i.e. airlines, airports, policy makers, etc) in the air transportation network during disruption management. Thus, a private decentralized AI for airline disruption management models the interaction amongst functional roles (i.e. low-level actors) within high-level stakeholder organizations during schedule execution and irregular operations. To this effect, a private decentralized AI presents a credible avenue for a system of systems modeling architecture for the AOCC \citep{Keating2003,DeLaurentis2005a}. In complement, a consortium decentralized AI for airline disruption management represents a hybrid of the \textit{Blockchain for AI} and \textit{AI for Blockchain} platforms that models the interaction amongst multiple high-level stakeholders within the air transportation system during disruption management. As such, a consortium decentralized AI provides an appropriate means for modeling irregular operations in the air transportation network as a federated system of systems \citep{Sage2001}.  
    
    \item \textbf{Application}: The application of a decentralized AI platform for airline disruption management allows for the following enhancements to existing decision support systems:
    \begin{enumerate}
        \item \textit{Autonomic computing} - This ensures that intelligent agents, such as functional roles in the AOCC, are able to cope when subjected to heterogeneity at all verticals including data sources, data processing, and data storage in order to perceive their internal states and conduct specified actions accordingly. As such, DLT provides a medium to permanently track the interactions  among multiple intelligent agents across separate levels of hierarchy within the air transportation system. 
        
        \item \textit{Optimization} - Existing paradigms for airline disruption management are enforced with centralized optimization strategies that impose global (i.e. system-wide) objectives such as maximizing airline profit margin. These optimization strategies result in strictly constrained and subordinate system-wide behaviors and performance to insure local objectives (such as minimizing flight delay duration) of constituent actors \citep{Clarke1998, Petersen2012}. Thus, the application of decentralized AI strategies through DLT (i.e. \textit{Blockchain for AI}) creates a distributed optimization platform that concurrently increases the search and design space to allow for improved system-wide performance for constrained and unconstrained interaction of intelligent agents in the AOCC.
        
        \item \textit{Planning} - From an airline stakeholder's  perspective, the outcome of the integration of AI and DLT during schedule execution is to readily attain a dynamic schedule planning scheme while managing airline disruptions. Existing integrated recovery paradigms are limited to static and tedious re-planning routines that take considerable amount of time (due to significant human interventions) to obtain an updated planning scheme. In that regard, \textit{Blockchain for AI} provides an applicable platform to obtain new planning strategies that utilize decentralized and distributed optimization routines for recording multiple planning schemes and their respective evolution histories (i.e. provenance) \citep{Salah2019}.  
        \item \textit{Knowledge discovery and management} - The management of data for separate intelligent agents in existing systems is centralized, and as such, creates a monolithic system that can not enable peer-to-peer interaction amongst system constituents for disruption management. A decentralized \textit{Blockchain for AI} platform provides a means for unstructured peer-to-peer network interaction of low-level stakeholders that represent functional roles or domains in the AOCC of an airline. Hence, emergent behavior in form of new sequences of resolution activities amongst multiple functional roles can be discovered during disruption management. In complement, \textit{AI for Blockchain} provides a medium for a structured peer-to-peer network interaction of high-level stakeholders (created via \textit{Blockchain for AI} platforms) that represent primary actors in the air transportation network, such as multiple airlines and airports. As such, emergent behavior in terms of different collaborative decision-making \citep{Ball2001a,Fearing2011} routines for airline disruption management can be enabled through autonomic AI computing.   
        
        \item \textit{Learning} - Existing AI platforms for airline disruption management are strictly trained and employed by using a centralized framework to achieve global intelligence. However, decentralized AI platforms are capable of autonomous and immutable learning through distributed computing that promotes fully coordinated local intelligence to achieve new  calibration routines for global intelligence. For instance, the sequence of real-time events to replan a disrupted airline network may require the interaction of many functional roles (i.e. intelligent agents) in the AOCC. As such, the transactions (i.e. interaction) amongst functional roles seeking consensus on a distributed ledger creates a platform to discover new learning routines for different scenarios of airline disruption.   
    \end{enumerate}
    \item \textbf{Operation}: The operation of a decentralized AI platform for airline disruption management represents its capacity to readily manage copious amounts of data for adaptable decision-making through AI applications. Unlike centralized AI systems with strict system specifications and operations, decentralized AI operations ensure versatility in data for separate functional roles in the AOCC through the following means:
    \begin{enumerate}
        \item \textit{Data management}: There is a high proclivity for data duplication in existing decision support systems because small changes in data content typically results in repeated (manual) transfer of updated datasets during disruption management \citep{Jarrah2000, Barnhart2003, Sherali2006}. As such, centralized data management becomes significantly inefficient as the airline route network (or air transportation network) expands. This inefficiency is characterized by rapid bandwidth overloading experienced by human specialists and increased backhaul network traffic that create substantial latency issues in existing decision support systems. Thus, decentralized data management provides an avenue for explicit decision support system modeling routines to be simultaneously developed and deployed at node levels that represent fundamental components of intelligent agents in a multi-agent system. Consequently, these decision support system modeling routines serve as dynamic metadata that can provide substantial latent information to a distributed ledger platform while tracking and maintaining provenance and immutability of disruption resolutions during consensus. 
        
        \item \textit{Learning model development} - Centralized learning creates a platform where learning models are calibrated and tested before software deployment \citep{Mosier2018}. As such, centralized platforms are unable to accommodate for rapidly evolving data streams because of the dynamic and cascading nature of the impact of airline disruptions on the air transportation network. In that regard, decentralized AI platforms, especially \textit{Blockchain for AI}, enable the premeditation and predetermination of all possible search rules for addressing a specific type of disruption from separate AI agents. These search rules are made readily accessible to a consensus algorithm based upon the engagement routines (i.e. smart contracts) that activate the instantiation of a specific search rule during negotiations amongst intelligent agents. 
        
        \item \textit{Model deployment} - Complementary to learning model development, AI replaces the brute force approach present in many centralized learning platforms that first solve an explicit optimization problem by trying every possible combination of decision variables, before recommending suitable solutions for the recovery of the airline network. Hence, DLT adopts different engagement routines to qualitatively and visually reveal the interaction of various search rules employed (i.e. learned and cached a priori from data) by multiple AI agents for real-time recovery of a disrupted airline network.
    \end{enumerate}
    
    \item \textbf{Consensus}: The consensus of a decentralized AI platform for airline disruption management relies on the capacity of multiple participating functional roles in the AOCC to agree on a set of acts for recovering a disrupted route network \citep{Bouarfa2018}. As such, the following artifacts provide pertinent properties for enforcing consensus in a decentralized multi-agent system for airline operations recovery. 
    \begin{enumerate}
        \item \textit{Byzantine fault tolerance (BFT)} - This represents a majority voting algorithm that eliminates transaction validation from malicious participants on a distributed ledger \citep{Lamport1982, Castro2002}. Malicious participants represent intelligent agents (or functional roles) in a permissioned DLT platform (i.e. multi-agent system) that can directly or indirectly manipulate the outcome of a recovery plan for a disrupted airline network. As such, many existing BFT algorithms guarantee consistent maximum fidelity and operability of a multi-agent system if at least two-thirds of all participants in the decentralized AI platform are not malicious \citep{Salah2019}. 
        
        \item \textit{Proof of stake (PoS)} - This represents a type of routine that selects stakeholders (e.g. functional roles in the AOCC) as signatories and validators of disruption events and corresponding disruption resolutions, respectively, during irregular operations \citep{Rauchs2018}. To that effect, stake in PoS indicates the degree to which the efficacy of the decisions (or resolutions) of a stakeholder can be trusted, or the measure of stakeholder interest in the effectiveness of a decentralized AI interaction for airline disruption management. A typical airline route network is somewhat delay-tolerant due to the finite capacity for useful resources in the air transportation network \citep{Lordan2016}. Hence, PoS provides an energy-efficient protocol for unconstrained interactions amongst stakeholders as long as stakeholders are defined by appropriately calibrated AI systems embedded with possible and readily applicable search rules for different disruption resolutions. 
        
        \item \textit{Smart contract} - This represents computational models (i.e. computer programs) that execute a set of engagements specified in digital form \citep{Magazzeni2017}. To this effect, constituent AI models for intelligent agents, such as functional roles in the AOCC, define the terms, conditions, and execution of a smart contract for airline disruption management. Furthermore, smart contracts assimilate and execute workflows, and as such, present a viable means for automating decision-making for regulatory compliance and approbation of several air transportation and flight operation processes during airline schedule execution. 
        \end{enumerate}
        
    \item \textbf{Infrastructure}: The infrastructure of a decentralized AI platform for airline disruption management represents the architecture (or environment) in which the interaction amongst separate stakeholders acting on a distributed ledger is simulated for any scenario of irregular operations. Many DLTs such as the Bitcoin blockchain are based upon a linear infrastructure wherein blocks of data content are sequentially connected via hashing mechanisms for a single chain link to achieve immutability and transparency. As such, single-chained DLTs (i.e. linear blockchain infrastructures) do not scale up well and struggle to achieve acceptable real-time performance necessary for effective decentralized applications \citep{Bencic2018}. Furthermore, to enable concurrent and independent interactions of multiple intelligent agents in a multi-agent system, distinct single chains are necessary to record the individual transaction activities of each intelligent agent, thereby rendering digital exchanges of assets, value, and information impossible for linear blockchain infrastructures in heterogeneous environments. To this end, nonlinear DLT infrastructures (based upon graph theory) provide a fitting platform for developing and deploying decentralized AI frameworks for airline disruption management that promote scalable real-time applications while utilizing big data.  
    
    Many existing nonlinear DLT infrastructures exist in the form of a directed acyclic graph \citep{Vanderweele2007,Karrer2009,Popov2018} that create multi-chain architectures such as parent-child chains and parallel chains. Parent-child chains represent multi-chain architectures where one or more chains serve as the primary chain that records the information and digital activities of other chains. As such, from a high-level perspective, a parent-child chain presents an appropriate nonlinear infrastructure for interaction in a consortium decentralized AI for airline disruption management, such that the primary chain is representative of a major policy maker (i.e. FAA or EASA) in the air transportation system while other chains in the consortium represent stakeholders like airlines and airports that are subordinate to the compliance directives (defined by smart contracts) provided by the policy maker. Parallel chains represent multi-chain architectures where constituent chains operate independently from other chains. Thus, the conscription of the rules of engagement for performing a transaction on a specific chain is controlled by a particular intelligent agent, and is sovereign from the actions of other intelligent agents operating on other chains in the decentralized platform. Hence, from a low-level perspective, a parallel multi-chain architecture provides a fitting nonlinear infrastructure for interaction in a private decentralized AI platform for airline disruption management, wherein each chain is representative of queuing disruption and resolution information from a particular functional role in the AOCC during airline schedule execution.   
\end{itemize}

\section{A Case Study} \label{case_study}

For an exhibition of the fusion of AI and DLT for airline disruption management, we consider a system of systems modeling of the network operations control center at Southwest Airlines (i.e. SWA-NOC).
\subsection{Methodology}
\begin{figure}[ht!]
	\centering
	\includegraphics[width=0.99\textwidth]{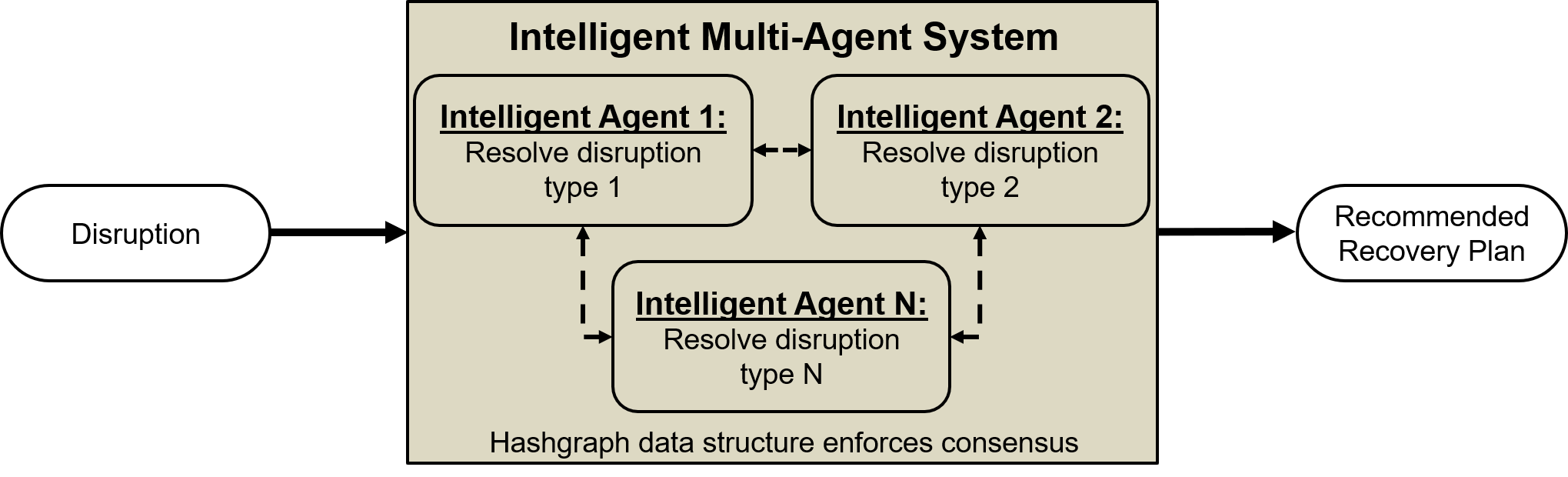}
	\caption{Intelligent multi-agent system for simultaneously-integrated recovery during airline disruption management}
	\label{fig:prop_paradigm}
\end{figure}

\begin{figure}[hb!]
	\centering
	\includegraphics[width=0.70\textwidth]{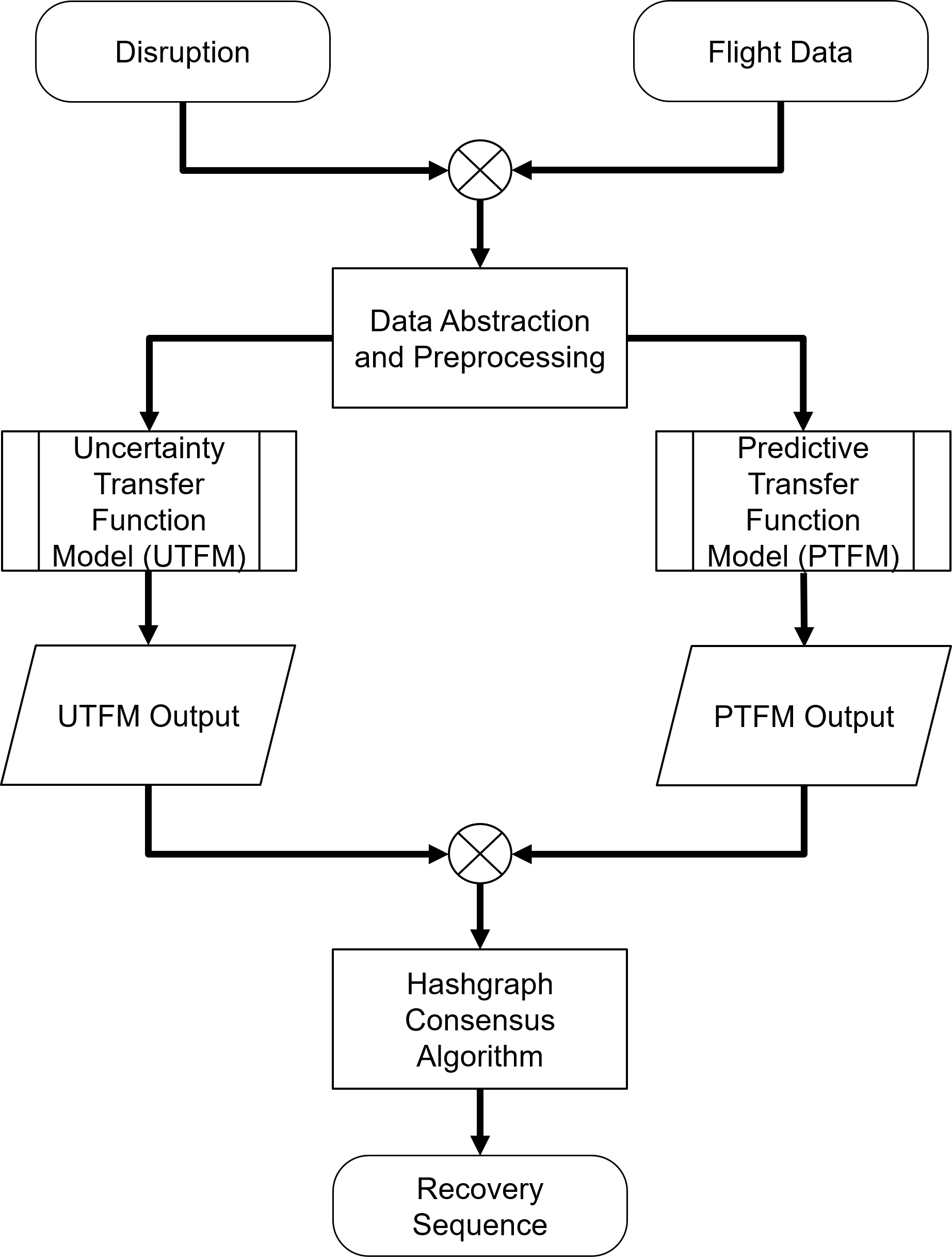}
	\caption{Generation of a recovery plan by a representative intelligent agent in the intelligent multi-agent system}
	\label{fig:IA}
\end{figure}

For our demonstration, each of the eleven functional roles in SWA-NOC is modeled as a separate AI system (i.e. intelligent agent) described by the intelligent multi-agent system schema shown in Fig.~\ref{fig:prop_paradigm}. Fig.~\ref{fig:prop_paradigm} reveals an original automatic (or semi-automatic) system for airline disruption management wherein functional parts of the AOCC (i.e. SWA-NOC) necessary for concurrent decision-making across all problem dimensions are characterized by an automated and intelligent multi-agent system framework that predominantly employs predictive analytics and blockchain mechanism design. The light brown box in Fig.~\ref{fig:prop_paradigm} defines the automated property of the overall system where there is a simultaneous interaction among different intelligent agents that represent $N$ separate managers for aircraft, crew, and passenger problem dimensions. To that effect, the intelligent agents are defined by data-driven and predictive decision-making AI models, and their interaction is defined by a consensus protocol (shown in \ref{alg:consensus}) for the Hashgraph data structure. 

Fig.~\ref{fig:IA} describes the manner in which a characteristic intelligent agent, shown in Fig.~\ref{fig:prop_paradigm}, develops a resolution sequence (i.e. recovery plan) for a certain type of disruption in original airline schedule. The first step is to obtain scheduled flight data and information on the corresponding type of disruption related to the problem dimension. Next, disruption information and flight schedule data are combined and transformed for use by AI models through data abstraction and preprocessing techniques \citep{Ogunsina2021}. Following data refinement, AI models defined by the uncertainty transfer function model (UTFM) and the predictive transfer function model (PTFM) are invoked, as extensively described by \cite{Ogunsina2021a} and \cite{Ogunsina2021b} respectively.
The UTFM provides the most likely sequence and likelihood of decision-making activities at different phases of flight for a specific resolution during disruption management. Thus, from an information theory perspective \citep{Jaynes1957, Cover2005}, the UTFM estimates the maximum amount of information or recovery uncertainty (measured as negative log likelihood) that can be predicted for a specific disruption resolution considered by a human specialist in a functional role during airline operations recovery. The recovery uncertainty estimated by the UTFM represents the reliability of the disruption resolution provided by the human specialist, which is subsequently used to define the stake of a specific functional role during its interaction with other functional roles in SWA-NOC to achieve consensus during disruption management. In complement, the PTFM provides an evaluation of the required duration and delay at different phases of flight upon enacting a particular resolution during disruption management. As such, from an airline operations recovery perspective \citep{Bratu2005,Medard2007,Petersen2012,Maher2015}, the PTFM estimates the recovery impact for a specific disruption resolution conceived by a human specialist in a functional role during airline disruption management.
\begin{figure}[ht!]
	\centering
	\includegraphics[width=0.99\textwidth]{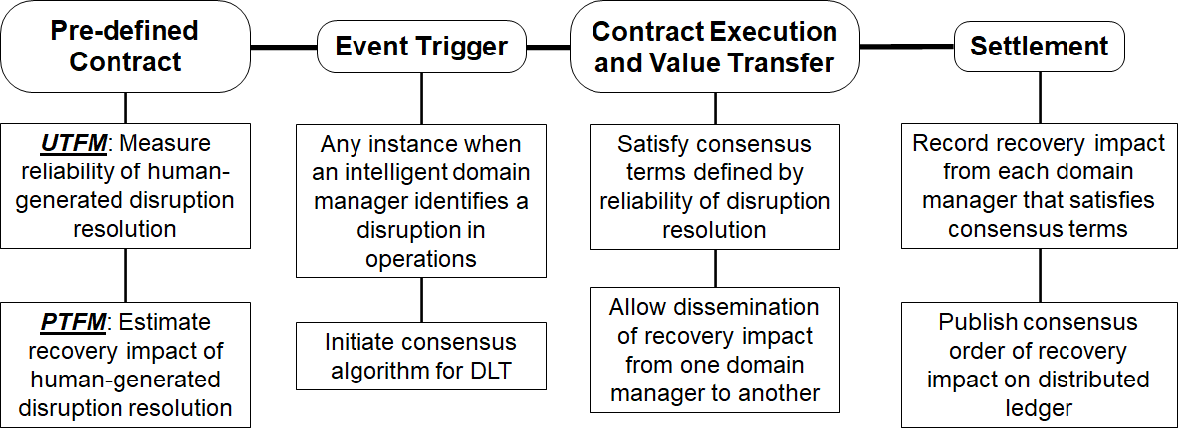}
	\caption{Integration and interaction routine in a decentralized AI platform for airline disruption management}
	\label{fig:IIR}
\end{figure}
Thus, for any functional role (i.e. intelligent agent or domain manager) operating based upon the consensus routine detailed in Fig.~\ref{fig:IIR} for a private decentralized AI platform, the UTFM and PTFM conjointly define a smart contract that measures the reliability of a human-generated disruption resolution and its corresponding effect on the recovery plan for a disrupted airline route network, thereby enabling integration in a multi-agent system for airline disruption management. Furthermore, the execution of the smart contract on a DLT is instantiated by the occurrence of disruptions that induce irregular operations for specific functional roles in SWA-NOC, thereby enabling interaction in a multi-agent system for airline disruption management. 

Hence, the multi-agent system for airline disruption management investigated in this case study represents a hybrid decentralized AI platform that enables multiple separate properties of the \textit{AI for Blockchain} and \textit{Blockchain for AI} platforms for integration and interaction respectively. To this end, the subsequent part of this section describes relevant rationale used to calibrate AI models as intelligent agents for integration into an applicable DLT protocol, which enables the creation and observation of interactions amongst separate intelligent agents during airline disruption management.  

\subsection{AI Calibration}
AI calibration is the process of adapting the behavior of an intelligent agent in a multi-agent system for a specific research scope and environment in order to attain high fidelity real-world simulations \citep{Murakawa2003,Ngo2012}. As such, to foster trust in airline network recovery obtained through human-AI collaboration, the estimation of the reliability of human-generated disruption resolutions by an intelligent agent during airline disruption management must be acceptable \citep{Lewandowsky2000,Okamura2020}. To this effect, we discuss the rationale used to calibrate the UTFM for intelligent agents that represent functional roles in SWA-NOC. 

\subsubsection{Definition}
The UTFM, detailed by \cite{Ogunsina2021a}, represents a design platform for creating probabilistic finite state machines \citep{Vidal2005a,Ghahramani2015} that define the manner in which disruption resolution activities, i.e. a series of input criteria provided by a human specialist, are translated by employing probabilities to evaluate the chance with which the disruption resolution yields the outcome intended by the human specialist. Hence, for a human specialist capable of providing a set of possible disruption resolutions (i.e. outcomes) ${r_{1}, ..., r_{k}}$, let $X$ represent a random variable that represents the unpredictable outcome of a single disruption resolution trial. As such, there exist values $p_{1}, ..., p_{k} \in [0, 1]$ wherein
\begin{equation}\label{eqn:prob1} 
    \begin{aligned}
        p_{i} = Pr[X = r_{i}]
    \end{aligned}
\end{equation}
and
\begin{equation}\label{eqn:prob2} 
    \begin{aligned}
        \sum_{i=1}^{k}p_{i} = 1
    \end{aligned}
\end{equation}
Note that the term $Pr[*]$ means `` the probability or likelihood of $*$ ". 

Thus, for every state $s \in S$ in a probabilistic finite state machine and every action or symbol $a \in A$ defined by a specific disruption resolution, the following expressions are valid: 
\begin{equation}\label{eqn:fsm} 
    \begin{aligned}
        either \quad \Delta(s, a) &= \emptyset \\
        \newline
        or \quad  \sum_{s^{'}\in \Delta(s, a)} Pr[s\xrightarrow{\text{a}} s^{\prime}] &= 1
    \end{aligned}
\end{equation}
Hence, for any series of input criteria $x \in A^{*}$ of length $k$ and every trace $t$ that defines the probabilistic finite state machine in the form:
\begin{equation}\label{eqn:tracet} 
    \begin{aligned}
        t = s_{0}, x_{1}, s_{1}, ... ,s_{k-1}, x_{k},s_{k}
    \end{aligned}
\end{equation}
the probability that the trace defined by Eqn~\ref{eqn:tracet} represents the path taken through the finite state machine upon reading the input criteria $x$ is expressed as:
\begin{equation}\label{eqn:probt} 
    \begin{aligned}
        Pr[t] = Pr[s_{0}\xrightarrow{x_{1}} s_{1}]\quad ...\quad Pr[s_{k-1}\xrightarrow{x_{k}} s_{k}]
    \end{aligned}
\end{equation}
Congruently, for all scenarios where $s^{\prime} \notin \Delta(s,a)$,
\begin{equation}\label{eqn:probt1} 
    \begin{aligned}
       Pr[s\xrightarrow{a} s^{\prime}] = 0
    \end{aligned}
\end{equation}
To this end, the probability that a disruption resolution provided by a human specialist is accepted by an aggregate probabilistic finite state machine $F$ that represents the UTFM, is based upon a set of traces $T(x)$ for $x$ wherein the final state $s_{k}$ of each trace is an accept state and $Pr[t]$ is summed over all such traces $t$ by the following expression: 
\begin{equation}\label{eqn:tracetsum} 
    \begin{aligned}
       Pr[x \quad \text{is accepted}] = \sum_{t\in T(x), s_{k}\in F} Pr[t]
    \end{aligned}
\end{equation}

\subsubsection{Learning Probabilities for Integration in SWA-NOC}
As discussed by \cite{Ogunsina2021a}, the real world application of probabilistic finite state machines for airline disruption management is made possible through a special state machine variant called the hidden Markov model. As such, the UTFM is representative of a hidden Markov model (i.e. a transducer-style probabilistic finite state machine) wherein outcomes are randomly generated at a state (in lieu of along transitions) according to a random stochastic process that models the activities at that state \citep{Allan2005}. Thus, for each state $s \in Q$ of the hidden Markov model, there exists a random variable associated with $s$ that takes in values of disruption resolution $x$ according to certain state-dependent probabilities. As such, it is imperative that these state-dependent probabilities are appropriately representative of the frequency of the random process (or variable) associated with a particular state instantiated during disruption management in the real world. To this effect, inferring the optimum structure (i.e. state transition and observation probabilities) of a hidden Markov model from real world data presents a common learning problem in AI.       

A variant of the expectation maximization algorithm \citep{Moon1996, Do2008}, called the Baum-Welch algorithm \citep{Munro2011}, provides a credible platform for learning the probabilities of a hidden Markov model from data through an unsupervised optimization routine that guarantees local convergence but not global convergence. As such, the Baum-Welch algorithm seeks to facilitate optimal prediction on training data that represents empirical frequency. However, empirical frequency does not provide a good estimate of the probability of new real-world situations. And just because an intelligent agent, defined by a model based upon empirical frequency, has not observed some value of a variable does not mean that the likelihood that the value exists should be zero (i.e., an impossible value does not exist) \citep{Poole2010}. Thus, to ensure high quality and fidelity of constituent UTFM structures for an intelligent agent, we employ prior domain knowledge on airline scheduling and operations to solve the zero-probability problem by implementing pseudocount (or prior count) for each value to which real-world data values are added during Baum-Welch training. 

To understand the pseudocount effect, suppose there is a binary feature $Y$ and a training data set with $n_{0}$ observed instances where $Y=0$ and $n_{1}$ instances where $Y=1$, then a pseudocount $c_{0} \geq 0$ is defined for $Y=0$ and another pseudocount $c_{1} \geq 0$ for $Y=1$ so as to estimate the probability (or inherent likelihood) of an outcome as:
\begin{equation}\label{eqn:pcount} 
    \begin{aligned}
       Pr[Y = 1] = \frac{n_{1}+c_{1}}{n_{0}+c_{0}+n_{1}+c_{1}}
    \end{aligned}
\end{equation}

Without loss of specificity, suppose $Y$ is defined by the domain ${y_{1}, ... , y_{k}}$, then a pseudocount $c_{i}$ is defined for each $y_{i}$ such that all prior counts are selected before training commences. As such, given a training data set with observed real-world examples where $n_{i}$ represents the number of instances with $Y = y_{i}$, then:
\begin{equation}\label{eqn:pcount1} 
    \begin{aligned}
       Pr[Y = y_{i}] = \frac{n_{i}+c_{i}}{\sum_{i} n_{i}+c_{i}}
    \end{aligned}
\end{equation}

Thus, choosing pseudocounts represents a supplementary part of designing the UTFM learner that effectively estimates how much an intelligent agent should accept (or believe) an input criterion $y_{i}$ if it had seen one instance with $y_{i}$ to be true during Baum-Welch training as compared to if it had seen no instances with $y_{i}$ to be true. As such, if there were no instances of $y_{i}$ to be true during training, then the intelligent agent believes that $y_{i}$ is impossible and $c_{i}$ was set as zero during training. However, setting $c_{i}$ as zero during Baum-Welch training of data sets with substantially small $n_{i}$ may cause the optimization to diverge, and not arrive at a local optimum due to the undefined nature of the logarithm for a value of zero. Therefore, prior counts provide a means to avoid to this problem. If there are instances with $y_{i}$ to be true during training, then the intelligent agent accepts that a value for $y_{i}$ observed one time would be $\frac{1+c_{i}}{c_{i}}$ times more likely than a value observed zero times. Thus, the prior knowledge factor, $f_{pk}$, for the consideration of a specific value in a data set during Baum-Welch training of constituent probabilistic finite state machines for the UTFM is expressed as:
\begin{equation}\label{eqn:pkf} 
    \begin{aligned}
       f_{pk} = \frac{1+c_{i}}{c_{i}} 
    \end{aligned}
\end{equation}

\begin{figure}[ht!]
	\centering
	\includegraphics[width=0.99\textwidth]{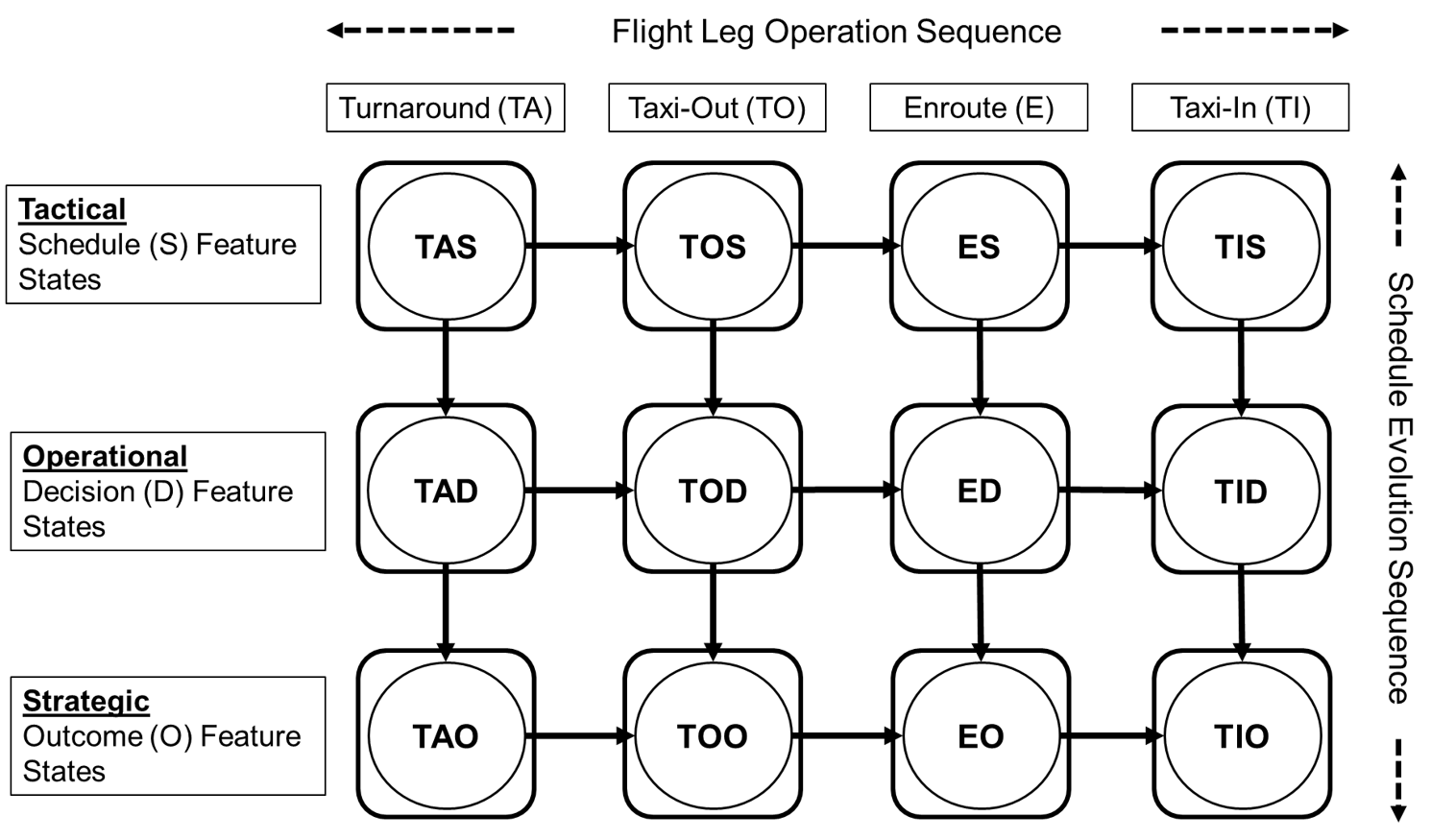}
	\caption{Phases of airline disruption management represented by UTFM \citep{Ogunsina2021a}}
	\label{fig:ADM_phases}
\end{figure}

For example, we use the binary feature that indicates a flight swap from the \textit{non-disrupted} data set to fix pseudocounts for Baum-Welch training of constituent probabilistic finite state machines for the UTFM, which define the random processes for remaining in the tactical and strategic phases of activity (shown in Fig.~\ref{fig:ADM_phases}) during disruption management. The \textit{non-disrupted} data set represents approximately 620,000 instances of Southwest Airlines flight schedules that executed without disruptions between September 2016 and September 2017. About 16,160 flight schedules in the \textit{non-disrupted} data set executed with a flight swap. The subset of flight schedules that executed with a flight swap were neither driven by the occurrence of active disruptions nor required irregular operations (i.e. considerable AOCC intervention) for execution. Theoretically, from an airline disruption management perspective, swapping a flight prior to schedule execution that is not necessitated by the existence of a disruption should be impossible. As such, the total number of instances of these flight schedules in the \textit{non-disrupted} data set represents an appropriate pseudocount value for eliminating the zero-probability problem during Baum-Welch training of tactical and strategic probabilistic finite state machines for the UTFM. Thus, there is at least a 2.61\% probability (i.e. inherent likelihood) that any value for remaining in the tactical and strategic phases of activity that is observed once would be 0.0062\% more likely than any value that is not observed at all during Baum-Welch training.  

\begin{sidewaystable}
\begin{center}
\caption{Pseudocount and inherent likelihood of observing new values for all functional roles in Southwest Airlines network operations control} \label{tab:pcount_summaries}
\begin{tabularx}{\linewidth}{ 
  | >{\centering\arraybackslash} p{0.2\textwidth}
  | >{\centering\arraybackslash} X
  | >{\centering\arraybackslash} X
  | >{\centering\arraybackslash} X
  | >{\centering\arraybackslash} X
  | >{\centering\arraybackslash} X
  | >{\centering\arraybackslash} X | }
  \hline
	\textbf{Functional Domain} & \textbf{Tactical Pseudocount} & \textbf{Operational Pseudocount} &\textbf{Strategic Pseudocount} & \textbf{Inherent Tactical Likelihood} & \textbf{Inherent Operational Likelihood} & \textbf{Inherent Strategic Likelihood} \\ \hline
	\textbf{\textit{Customer Hold}} & 1,906 & 11,289 & 3,222 & 0.041 & 0.243 & 0.069 \\ \hline
	\textbf{\textit{Dispatch CSC}} & 1,988  & 3,160 & 4,792 & 0.115 & 0.183  & 0.277 \\
	\hline
	\textbf{\textit{Flight Operations}} & 6,365  & 10,580 & 2,682 & 0.177 & 0.294 & 0.074\\
	\hline
	\textbf{\textit{Fuel Management}} & 603 & 1,180  & 228 & 0.126 & 0.246 & 0.048\\
	\hline
	\textbf{\textit{Ground Operations}} & 8,146 & 48,827 & 5,748  & 0.049 & 0.293 & 0.034\\ \hline
	\textbf{\textit{Inflight}} & 4,751  & 25,423 & 3,505 & 0.060 & 0.323 & 0.045 \\
	\hline
	\textbf{\textit{Maintenance}} & 6,985 & 4,901 & 4,648 & 0.211 & 0.148 & 0.140\\
	\hline
	\textbf{\textit{NAS}} & 2,221 & 2,774 & 1,184 & 0.099 & 0.124 & 0.053\\
	\hline
	\textbf{\textit{Security}} & 850 & 955 & 145 & 0.291 & 0.326 & 0.050 \\
	\hline
	\textbf{\textit{Technology}} & 869 & 2,206  & 397 & 0.098 & 0.249 & 0.045\\
	\hline
    \textbf{\textit{Weather}} & 1,483 & 597 & 1,065 & 0.118 & 0.048 & 0.085\\
    \hline
\end{tabularx}
\end{center}
\end{sidewaystable}

The arrival of a flight at the exact time planned prior to schedule execution is almost impossible when there are disruptions in the airline route network \citep{Ball2010}. As such, we use the binary feature that indicates precise on-time arrival (i.e. $A0$) from the \textit{disrupted} data set to set pseudocounts for Baum-Welch training of probabilistic finite state machines for the UTFM that define the random processes for remaining in the operational phase of activity during airline disruption management. The \textit{disrupted} data set contains about 434,000 instances of Southwest Airlines flight schedules that were subject to flight delays across multiple functional roles in the AOCC between September 2016 and September 2017. In that regard, we use the binary features that indicate route origination (i.e. first scheduled flight on day of operation) and flight swap from the \textit{disrupted} data set to specify pseudocounts for Baum-Welch training of probabilistic finite state machines that define the transitions along and across different phases of activity in the UTFM respectively. 

Table~\ref{tab:pcount_summaries} reveals the pseudocount and inherent likelihood of observing new values (i.e. disruption resolution criteria) for each functional role in Southwest Airlines network operations control, based upon an assessment of a training subset of the \textit{disrupted} data set that represents 99\% of all instances of flight schedule execution in the \textit{disrupted} data set. Note that the remaining 1\% represent instances of flight schedules in the \textit{disrupted} data set that are used later in this paper as new (disrupted) flight schedules for demonstrating interaction in the decentralized AI platform. As shown in Table~\ref{tab:pcount_summaries}, the Security functional role has the highest inherent likelihoods of 29.1\% and 32.6\% for observing new values for tactical and operational disruption management respectively. Similarly, the Dispatch CSC functional role has the highest inherent likelihood of 27.7\% for observing a new value for strategic disruption management. Conversely, the lowest inherent likelihoods of observing a new value for tactical, operational, and strategic disruption management are 4.1\%, 4.8\%, and 3.4\% respectively, and are realized by the Customer Hold, Weather, and Ground Operations functional roles. Thus, human specialists in the Security and Dispatch CSC functional roles have the most trial-and-error learning tendencies during different phases of airline disruption management. In contrast, human specialists in the Weather, Customer Hold, and Ground Operations functional roles have the least trial-and-error learning proclivities during separate phases of airline disruption management.   

\subsection{DLT Protocol}
DLT presents a means to rive and improve airline scheduling and operations \citep{Salah2019}. To this effect, we employ a special type of nonlinear parallel-chain DLT called Hashgraph, which offers four pertinent advantages for achieving adaptive and dynamic decision-making interactions that are not readily available in current decision support systems for airline operations recovery and disruption management \citep{Baird2018a}. 

\begin{enumerate}
    \item \textbf{Performance} - The Hashgraph consensus algorithm can process hundreds of thousands of transactions per second, and thus affords an almost-perfect efficiency in bandwidth usage for a fully connected peer-to-peer network of intelligent agents that represent functional roles in SWA-NOC. This ensures that consensus latency (i.e. real time recovery negotiation amongst intelligent agents) is restricted to seconds, as opposed to minutes observed in existing approaches for simultaneously-integrated recovery in airline disruption management \citep{Castro2014}. 
    
    \item \textbf{Security} - Existing platforms for simultaneously-integrated recovery in airline disruption management employ consensus methods that require coordinators, leaders and communication downtime to enable interaction amongst agents in a multi-agent system \citep{Bouarfa2018}. As such, these platforms are vulnerable to distributed denial of service (DDoS) \citep{Lau2000,Zargar2013} attacks that compromise the integrity of the multi-agent system by allowing multiple systems to innudate the bandwidth and resources of a targeted system. In that regard, Hashgraph is immune to DDoS attacks against the consensus algorithm because there are no leaders or coordinators in the data infrastructure. As such, Hashgraph enables the ultimate benchmark for security currently attainable in distributed consensus through asynchronous byzantine fault tolerance (ABFT) \citep{Castro2002,Hao2018,Baird2018a}. ABFT ensures that the consensus order of transactions (i.e. disruption resolutions) match the actual order in which the transactions posted to the distributed ledger are resolved by intelligent agents. To that effect, it is almost impossible for a single intelligent agent to prevent the writing of transaction information to the distributed ledger, or influence the order of transactions from consequent consensus in the multi-agent system. 
    
    \item \textbf{Governance} - The decentralized AI network hosted on the Hashgraph platform is governed concurrently by all eleven functional roles in the AOCC that represent intelligent agents in a multi-agent system, wherein no intelligent agent has control and no small group of intelligent agents has exorbitant influence over the multi-agent system as a whole. This eliminates the need for centralized control observed in many existing monolithic systems for airline disruption management.  
    
    \item \textbf{Stability} - Hashgraph presents a technology and platform that ensures that intelligent agents automatically validate the ancestry of the information circulated on the distributed ledger prior to deployment through a shared state mechanism. As such, the state mechanism on the Hashgraph platform for airline disruption management is defined by the UTFM and PTFM, which represent unilateral technical controllers in the decentralized AI network during schedule recovery. Furthermore, Hashgraph enables human specialists or supervisors (i.e. platform and software developers) in the AOCC to specify changes and updates to components of intelligent agents, such that the updates are automatically adopted for all intelligent agents at precisely the same time. This ensures that any intelligent agent with antiquated updates is unable to modify the distributed ledger or tender its version of the ledger as valid.
    
\end{enumerate}

\subsubsection{Interaction in SWA-NOC as a Byzantine Generals Problem}
Solving the Byzantine Generals Problem \citep{Lamport1982} represents one of the most challenging methods for verifying and validating the reliability of a computerized multi-agent system in order to manage the failure of one or more of its constituent agents. Conceptually, the Byzantine Generals Problem describes the communication of separate generals in several divisions of the Byzantine army camped outside an enemy city, with only one messenger available to each general. The objective of the generals' mission is to decide on the best collective plan of action after each general observes the behavior (i.e. stratagem) of the enemy. However, some generals may be betrayers that try to hinder loyal generals from achieving a consensus on the best plan of action to defeat the enemy. Thus, solving the Byzantine Generals Problem is analogous to enabling interaction amongst functional roles in SWA-NOC during disruption management. In this analog, each intelligent agent for a functional role represents a Byzantine general and the enemy represents a disruption in schedule and operations at a particular airline station. The collective plan of action (i.e. recovery plan) describes the order in which disruption resolutions from multiple functional roles are implemented. As such, the messenger for an intelligent agent is defined by cryptographic keys \citep{Blakley1979} that encode disruption resolutions from appropriate AI models imbued in the intelligent agent. In that regard, the objective of the interaction amongst intelligent agents in a multi-agent system for airline disruption management is to agree on the best recovery plan of action for a disrupted airline schedule such that the following conditions are upheld: 
\begin{itemize}
    \item All intelligent agents representative of separate functional roles in SWA-NOC decide upon the same recovery plan of action.  
    
    \item A small number of traitors (i.e. intelligent agents) cannot cause loyal intelligent agents to employ a bad recovery plan.
\end{itemize}

Thus, the intelligent agents operating in a decentralized multi-agent system must have a robust algorithm to guarantee these conditions. To that effect, the Swirlds (Hashgraph) consensus algorithm provides an appropriate medium for insuring that the aforementioned conditions are always satisfied during disruption management and operations recovery. 

\subsubsection{Hashgraph for Consensus in SWA-NOC}
The algorithms that solve the Byzantine agreement problem typically exchange many messages for intelligent agents to vote. For many multi-agent systems, a single YES/NO decision by $n$ intelligent agents can require up to $\mathrm{O}(n^{3})$ messages to be sent across the network \citep{Correia2011, Castro2014}. And extending an algorithm to decide a total order on a set of transactions for a single YES/NO decision can further increase the voting traffic and compound latency problems. However, the Swirlds consensus algorithm \citep{Baird2018a}, which addresses Byzantine agreement on the Hashgraph DLT, employs a virtual voting mechanism that sends zero votes over the network for a multi-agent system. Thus, we define pertinent concepts that enable the applicability of the Swirlds (Hashgraph) consensus algorithm to the decentralized AI network for airline disruption management as follows \citep{Baird2018, Baird2020}:

\begin{itemize}
    \item \textit{Transactions}: A transaction occurs when any intelligent agent representative of a functional role in SWA-NOC publishes its disruption resolution data and corresponding timestamp (i.e. time when the data is appended) to the distributed Hashgraph ledger. The disruption resolution data contains two pieces of information namely: 
    \begin{enumerate}
        \item The recovery uncertainty or reliability measure (i.e. entropy) of the set of actions for disruption resolution generated by the UTFM in the intelligent agent.
        \item The recovery impact measured in terms of turnaround duration, block time, tactical delay, and strategic delay estimated by the PTFM in the intelligent agent. Tactical delay and strategic delay represent discretionary holdup applied by a human specialist prior to flight schedule execution.
    \end{enumerate} 
    Any intelligent agent in the multi-agent system (i.e. decentralized AI network) can create a signed transaction at any time when a disruption in scheduled airline operations occurs. All intelligent agents in the multi-agent system receive a copy of the transaction, and the decentralized AI network reaches Byzantine agreement on the order of all transactions.
    
    \item \textit{Fairness}: This ensures that it is difficult for a small group of dubious intelligent agents to unfairly influence the order of transactions that is selected as consensus.
    
    \item \textit{Gossip}: This represents how disruption resolution information is disseminated by each intelligent agent repeatedly selecting another intelligent agent at random, and telling them all they know. 
    
    \item \textit{Hashgraph}: This represents the data structure or ledger upon which transaction records are gossiped and in what order they are gossiped.
    
    \item \textit{Gossip-about-gossip}: This indicates the gossip protocol that the \textit{Hashgraph} employs for its operation. The information being gossiped is the history of the gossip itself, and not the disruption resolution data contained in the gossip. As such, only a small amount of bandwidth is required for gossiping transactions amongst intelligent agents. 
    
    \item \textit{Virtual voting}: Every intelligent agent is privy to a copy of the \textit{Hashgraph}. For instance, an intelligent agent that represents the Security functional role can estimate the vote that another intelligent agent representative of the Ground Operations functional role would have sent if they both had been executing a traditional Byzantine agreement protocol that required them to send votes. As such, the intelligent agent representing Ground Operations does not need the intelligent agent representing the Security functional role to actually vote. To that effect, every intelligent agent can attain Byzantine agreement on any number of recovery plan actions without a single vote ever being sent. The \textit{Hashgraph} data structure alone is sufficient, so no additional bandwidth is expended beyond that required for gossiping the data structure. 
    
    \item \textit{Famous witnesses}: There are situations during interaction in the multi-agent system where a list of $n$ transactions are ordered by running separate Byzantine agreement protocols on $\mathrm{O}(n \log n)$ different YES/NO questions that seek to answer whether a disruption resolution event $x$ occurred before another disruption resolution event $y$. To address these situations in an expedient manner, a few disruption resolution events (i.e. vertices in the \textit{Hashgraph}) called \textit{witnesses} are selected that define a witness to be \textit{famous} if the \textit{Hashgraph} data structure shows that majority of the participating intelligent agents on the ledger received a particular disruption resolution event fairly soon after it was created. Thus, it is sufficient to execute the Byzantine agreement protocol only for witnesses that decide if a particular witness is famous for each intelligent agent. As such, it becomes relatively easy to retrieve a fair total order for all disruption resolution events upon attaining Byzantine agreement on the exact set of famous witnesses. 
    
    \item \textit{Strongly seeing}: This property is defined as the ability of any two disruption resolution events (i.e. vertices) $x$ and $y$ in the \textit{Hashgraph} to be connected by multiple directed paths passing through enough participating intelligent agents on the distributed ledger. For instance, if the intelligent agents for the Security and Ground Operations functional roles are able to independently estimate the virtual vote on a specific question from the intelligent agent representing the Weather functional role, then the intelligent agents for the Security and Ground Operations functional roles will get the same answer. The proof of the lemma that demonstrates this proposition forms the foundation for the subsequent mathematical proof of Byzantine agreement with a probability of one, which is achieved by the Swirlds consensus algorithm \citep{pyswirld2017, Crary2021}.
    
    \item \textit{Asynchronous Byzantine Fault Tolerance}: The Byzantine fault tolerance theorem states that there is absolute certainty that each disruption resolution event $x$ created by an honest intelligent agent will eventually be assigned a consensus position in the total order of disruption events. Asynchrony for Byzantine fault tolerance is ensured based upon the assumption that the digital signatures and cryptographic hashes are secured, such that signatures can not be forged, signed messages can not be altered without being detected, and hash impingement can never be found \citep{Barker2012}. As such, syncing the gossip protocol ensures that when an intelligent agent $a$ sends intelligent agent $b$ all the disruption resolution data it knows, $b$ accepts only those that have a valid signature and contain valid hashes corresponding to the disruption resolution data that it has available. Therefore, the following definitions describe the property of asynchronous Byzantine fault tolerance in the intelligent multi-agent system for obtaining a recovery plan during airline disruption management:
    
    \begin{enumerate}
        \item A disruption resolution event $x$ is defined to be an ancestor of a disruption resolution event $y$ if $x$ is $y$, a parent of $y$, a parent of a parent of $y$ and so on. Furthermore, $x$ is also a self-ancestor of $y$ if $x$ is $y$, or a self-parent of $y$, or a self-parent of a self-parent of $y$ and so on.
        \item The \textit{round created number} (i.e. round) of a disruption resolution event $x$ is $r+i$, where $r$ is the maximum round number of the parents of $x$ or 1 if it has no parents, and $i$ is 1 if $x$ can strongly see more than $2n/3$ witnesses in round $r$ or 0 otherwise. Note that $n$ is the total number of participating intelligent agents.  
        \item The \textit{round received number} (i.e. received round) of a disruption resolution event $x$ is the first round where all unique famous witnesses are descendants of $x$.
        \item The pair of disruption resolution events $(x,y)$ is a fork (i.e. an intelligent agent update collectively agreed upon) if $x$ and $y$ have the same creator but neither of them is a self-ancestor of the other.
        \item An honest intelligent agent participant on the distributed ledger tries to sync infinitely often with every other participating intelligent agent, creates a valid disruption resolution event after each sync with cryptographic hashes of the latest self-parent and other parents, and will never create two disruption resolution events that are forks with each other. 
        \item A disruption resolution event $y$ can be seen by a disruption resolution event $x$ if $y$ is an ancestor of $x$, and the ancestors of $x$ do not include a fork by the creator of $y$.
        \item A disruption resolution event $x$ can \textit{strongly see} disruption resolution event $y$ if $x$ can see $y$ and there is a set of disruption resolution events $R$ from more than two-thirds of participant intelligent agents, such that $x$ can see every disruption resolution event in $R$ and $y$ can be seen by every disruption resolution event in $R$.
        \item A witness is the first disruption resolution event created by an intelligent agent in a round.
        \item A unique famous witness is a \textit{famous witness} that does not have the same creator as any other famous witness created in the same round.
        \item \textit{Hashgraph} $P$ for intelligent agent $a$ and \textit{Hashgraph} $Q$ for intelligent agent $b$ are said to be consistent (i.e. exactly the same) if and only if for any disruption resolution event $x$ in both \textit{Hashgraphs}, there are the same set of ancestors for $x$ with the same parent and self-parent edges between the set of ancestors. 
    \end{enumerate}

    \item \textit{Consensus}: The stake of each intelligent agent is defined by a positive integer that represents the total entropy (i.e. information) of the set of recovery activities generated by its UTFM. As such, the vote of an intelligent agent in the multi-agent system, while creating a recovery plan, is weighted proportional to its voting stake during each round of voting. Thus, consensus in the multi-agent system is defined by a set of intelligent agents whose combined voting stake is more than $2n/3$, where $n$ is the total stake of all participating intelligent agents. In complement, the consensus timestamps of disruption resolution events for a set of disruption resolution events $R$ is the median of the timestamps in $R$ weighted by the voting stake of the creator of each event in $R$. As such, the weighted median is analogous to selecting each disruption resolution event $x$ in $R$ and placing multiple copies of the timestamp of $x$ into a basket, such that the number of timestamp copies is equal to the stake of the intelligent agent that created $y$, and then choosing the median of the timestamps in the basket. These routines form the basis for proof of stake in achieving consensus in the multi-agent system.   
    
\end{itemize}

All mathematical proofs, lemmas, and algorithmic routines that enable all the aforementioned concepts for the Hashgraph consensus framework can be found in \cite{Baird2018} and \cite{Crary2021}.     

\subsubsection{Defining Stake for Functional Roles in SWA-NOC}
Recall from Fig.~\ref{fig:ADM_phases} that the UTFM is comprised of a total of 12 interconnected states for three separate phases of activity; that is, four schedule states for tactical disruption management, four decision states for operational disruption management, and four outcome states for strategic disruption management. Thus, in principle, the trace between any two states (or phases of activity) in the UTFM can be used to define the stake of a functional role. Furthermore, the stake of an intelligent agent operating on the Hashgraph consensus platform must be expressed as a positive integer \citep{Baird2018}. As such, we employ a method that measures the stake of a functional role in the AOCC as the total entropy of the trace (i.e. path) from the first schedule state (i.e. Turnaround Schedule or TAS) to the last outcome state (i.e. Taxi-In Outcome or TIO), given a set of input criteria for each phase of activity in the UTFM. Mathematically, the entropy $E$ of a discrete random variable $T$ representing the trace with probability mass function $t(x)$ is expressed as:

\begin{equation}\label{eqn:entropy} 
    \begin{aligned}
       E(T) = -\sum_{x} {t(x)\log_{2}{t(x)}}
    \end{aligned}
\end{equation}

\begin{table}[t!]
\caption{Position weights of states at separate phases of activity in the UTFM for an intelligent agent} \label{tab:position_weight}
\begin{tabularx}{\linewidth}{ 
  | >{\centering\arraybackslash}p{0.2\textwidth}
  | >{\centering\arraybackslash}X 
  | >{\centering\arraybackslash}X | }
  \hline
	\textbf{ADM Phase} & \textbf{Possible States} & \textbf{Position Weight}\\ \hline
	\textbf{\textit{Tactical}} & TAS, TOS, ES, TIS & 1\\ \hline
	\textbf{\textit{Operational}} & TAD, TOD, ED, TID & 4\\ \hline
	\textbf{\textit{Strategic}} & TAO, TOO, EO, TIO & 2\\ \hline
\end{tabularx}
\end{table}

Thus, the entropy is the expected number of bits necessary to communicate the value of the trace $T$ if the best possible method (i.e. coding scheme) is used to estimate the distribution of $t(x)$. As such, we estimate $t(x)$ as the probability of the most likely sequence of transitions (i.e. the Viterbi likelihood) from Turnaround Schedule to Taxi-In Outcome for a given set of input (or action) criteria $x$ provided by a human specialist in the AOCC. However, we can not guarantee that $t(x)$ represents the globally optimal distribution for communicating the value of trace $T$ because the best possible ``coding scheme" for $t(x)$ was obtained through Baum-Welch training, which only guarantees local optimality. Thus, we create a surrogate value $S$ of trace $T$ based upon domain knowledge that describes the significance of the value management of flight operation at separate phases of activity in airline disruption management. To estimate the surrogate value, we assign a position weight \citep{Sinha2006,Xia2012} to states at each phase of activity, such that the lowest position weight is assigned to states for tactical disruption management and the highest position weight is assigned to states for operational disruption management. States for tactical and strategic disruption management receive lower position weights, as compared to states for operational disruption management, because these states are only active prior to the execution of a flight schedule and do not rapidly diminish the real-time value of flight recovery.    Table~\ref{tab:position_weight} shows the position weights for states at distinct phases of activity in the UTFM for a functional role in the AOCC. Since entropy is measured in bits, we adopt a binary position weighting scheme for any trace from Turnaround Schedule to Taxi-In Outcome, such that each state observed for tactical and strategic disruption management is assigned a value of 1 and 2 respectively, while each state observed for operational disruption management is assigned a value of 4. As such, $S$ represents the sum of the position weights $s(x)$ for $n$ transitions from Turnaround Schedule to Taxi-In Outcome by following the Viterbi path (i.e. the most likely sequence of actions from Turnaround Schedule to Taxi-In Outcome obtained through Viterbi decoding). To this end, we amend the entropy estimated in Eqn.~\ref{eqn:entropy} to information cross entropy, which defines the expected number of bits necessary to communicate the value (or information) taken by $T$ for employing the plausibly sub-optimal Baum-Welch coding scheme defined by $t(x)$. Mathematically, the information cross entropy ($ICE$) is expressed as:

\begin{equation}\label{eqn:centropy} 
    \begin{aligned}
       ICE(s,t) = -S\log_{2}{t(x)}= -n\sum_{x} {s(x)\log_{2}{t(x)}}
    \end{aligned}
\end{equation}

Thus, the voting stake $v$ of a functional role in the AOCC on a single transaction during airline disruption management is expressed as:
\begin{equation}\label{eqn:stake} 
    \begin{aligned}
       v = \floor{ICE}
    \end{aligned}
\end{equation}

\section{Results and Discussion} \label{results}
This section describes the results from a demonstration of our decentralized AI platform for a multi-agent system that creates airline disruption management and schedule recovery plans, through the interaction of intelligent agents that perform transactions on the Hashgraph data structure and consensus platform. In that regard, the following assumptions, conditions and terms apply for our demonstration:

\begin{table}[ht!]
\caption{Total number of queued disrupted flight schedules for each functional role in SWA-NOC used for case study} \label{tab:queued_delay}
\begin{tabularx}{\linewidth}{ 
  | >{\centering\arraybackslash}p{0.3\textwidth}
  | >{\centering\arraybackslash}X 
  | >{\centering\arraybackslash}X | }
  \hline
	\textbf{Functional Role} & \textbf{Total Number of Queued Disrupted Flight Schedules} & \textbf{Affected Problem Dimension}\\ \hline
	\textbf{\textit{Customer Hold}} & 469 & Aircraft and Passenger\\ \hline
	\textbf{\textit{Dispatch CSC}} & 175 & Aircraft and Crew\\ \hline
	\textbf{\textit{Flight Operations}} & 364 & Crew\\ \hline
	\textbf{\textit{Fuel Management}} & 49 & Aircraft\\ \hline
	\textbf{\textit{Ground Operations}} & 1684 & Aircraft and Passenger\\ \hline
	\textbf{\textit{Inflight}} & 795 & Crew\\ \hline
	\textbf{\textit{Maintenance}} & 336 & Aircraft\\ \hline
	\textbf{\textit{NAS}} & 227 & All\\ \hline
	\textbf{\textit{Security}} & 30 & Passenger\\ \hline
	\textbf{\textit{Technology}} & 90 & All\\ \hline
	\textbf{\textit{Weather}} & 127 & All\\
	\hline
\end{tabularx}
\end{table}

\begin{enumerate}
    \item We use a data subset that represents 1\% of the \textit{disrupted} data set to create a disrupted airline route network, which involves multiple functional roles in SWA-NOC. This data subset represents disrupted flight schedules that executed sometime between September 2016 and September 2017 in the Southwest Airlines route network, and are not used for developing (i.e. calibrating or training) the UTFM and PTFM for intelligent agents.
    \item We assume that the disrupted flight schedules define all or part of a ``fictitious" and arbitrary route network served by Southwest Airlines in a representative day of the year. 
    \item A new transaction, created by an intelligent agent representative of a functional role in the AOCC, is stored as random bytes on the \textit{Hashgraph} ledger, based upon the queue number of a specific disruption in the irregular operations database of the functional role.
    \item We assume that all queued disrupted flight schedules in the database of the functional role for each intelligent agent have a set of corresponding disruption resolutions (i.e. input criteria) provided by human specialists in the AOCC. With respect to the \textit{disrupted} data set, these input criteria represent actual disruption resolutions used to recover the disrupted flight schedules that executed at a particular time, and serve as concurrent inputs for the UTFM and PTFM.
    
    \item We assume that all human specialists in the AOCC, who provide input criteria for the decentralized AI platform (or disruption resolutions for the multi-agent system), are flexible and expert decision-makers in their respective functional roles. Hence, they are capable of expediently proffering \textit{rules-of-thumb} to address disruptions that affect their respective functional roles.  
\end{enumerate}

Table~\ref{tab:queued_delay} shows the total amount of queued disrupted flight schedules for each functional role in the Southwest Airlines network operations control center. Southwest Airlines operates over $4,100$ flight schedules every day \citep{Deloitte2017}. As such, there are a total of $4,364$ disrupted flight schedules, which represent the worst plausible irregular operations scenario where all flight schedules in the arbitrary airline route network are disrupted at the same time on a representative day of the year. We employ the Python programming language for the amalgamation of the Hashgraph consensus algorithm and the AI models for the intelligent agents that represent functional roles in SWA-NOC. This enables the operation and deployment of the decentralized AI platform for airline disruption management.   

\subsection{Data Structure}

\begin{figure}[ht!]
	\centering
	\includegraphics[width=0.99\textwidth]{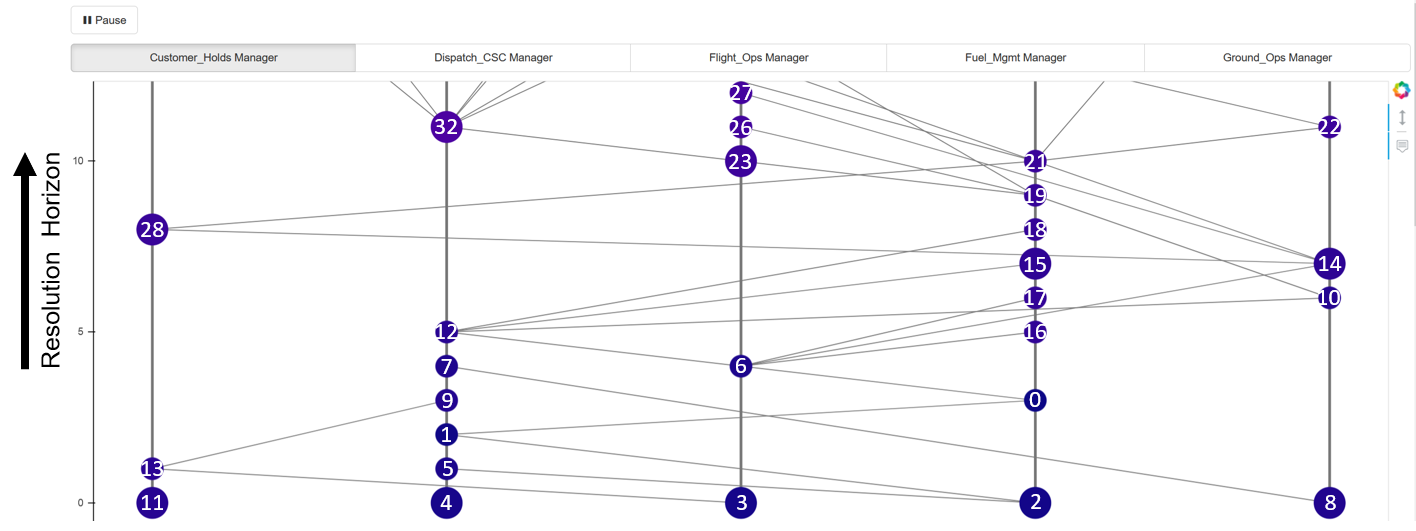}
	\caption{Hashgraph data structure revealing a recovery plan from a few functional roles in SWA-NOC}
	\label{fig:Hashgraph_rec}
\end{figure}

Fig.~\ref{fig:Hashgraph_rec} shows the \textit{Hashgraph} (i.e. distributed ledger data structure) for creating a simultaneously-integrated recovery plan that involves five of the eleven functional roles in SWA-NOC. We use five functional roles for this specific demonstration for ease of readability. As such, there are five participating intelligent agents (or domain managers) that represent the following functional roles: Customer Hold, Dispatch CSC, Flight Operations, Fuel Management, and Ground Operations, respectively. The circles shown in Fig.~\ref{fig:Hashgraph_rec} represent disruption resolution events, such that the larger circles are disruption resolution events that are also \textit{famous witnesses}. In addition, the dark blue circles towards the bottom of the ledger shown in Fig.~\ref{fig:Hashgraph_rec} represent disruption resolution events from the first round of consensus, while the purple circles located towards the top of the ledger represent resolution events from the second round of consensus.  Furthermore, the non-negative integers shown in Fig.~\ref{fig:Hashgraph_rec} represent the position (or place) of a specific disruption resolution event in the consensus order of disruption resolution events, which describe the recovery plan of action during disruption management by the five functional roles. The resolution horizon, which goes from bottom to top (i.e. moving upwards) as shown in Fig.~\ref{fig:Hashgraph_rec}, describes the manner in which a disruption resolution event or transaction is added to the distributed ledger (i.e., timestamp) by an intelligent agent. Hence, the grey lines represent the gossip paths that indicate how disruption events occur and how corresponding resolution information is disseminated amongst the intelligent domain managers. Therefore, the recovery plan of action agreed upon by the five functional roles, as shown in Fig.~\ref{fig:Hashgraph_rec}, is to apply the resolution for the second disruption encountered by the Fuel Management functional role first (i.e. small circle labeled $0$) before applying the resolution for the third disruption encountered by the Dispatch CSC functional role next (i.e. small circle labeled $1$), and so on and so forth. In that vein, the recovery plan of action represents a trace $r$ that follows a set of non-negative integers in ascending order, $q$, such that each integer in $q$ represents the label of a corresponding disruption resolution event, as shown in Fig.~\ref{fig:Hashgraph_rec}. Therefore, the recovery plan of action from all five functional roles is expressed as:

\begin{sidewaystable}
\begin{center}
\caption{Summaries of the first ten disruption resolution events from Hashgraph data structure shown in Fig.~\ref{fig:Hashgraph_rec}} \label{tab:hashgraph_summaries}
\begin{tabularx}{\linewidth}{ 
  | >{\centering\arraybackslash} X
  | >{\centering\arraybackslash} X
  | >{\centering\arraybackslash} X
  | >{\centering\arraybackslash} p{0.2\textwidth}
  | >{\centering\arraybackslash} X
  | >{\centering\arraybackslash} X
  | >{\centering\arraybackslash} X
  | >{\centering\arraybackslash} X | }
  \hline
    	\textbf{Consensus Position} & \textbf{Famous Witness} & \textbf{Arbitrary Flight ID} & \textbf{Functional Role} & \textbf{Estimated Tactical Delay Duration (mins)} &\textbf{Estimated Turnaround Duration (mins)} & \textbf{Estimated Block Time Duration (mins)} & \textbf{Estimated Strategic Delay Duration (mins)} \\ \hline
	0 & NO  & 1536  & \textit{Fuel Management} & 61 & 18 & 126 & 26 \\ \hline
	1 & NO & 3222 & \textit{Dispatch CSC} & 41 & 22 & 93 & 8 \\
	\hline
	 2 & YES & 2201 & \textit{Fuel Management} & 7 & 17  & 247 & 33 \\
	\hline
	3 & YES & 11670  & \textit{Flight Operations} & 35 & 1 & 106 & -14 \\
	\hline
	4 & YES & 12388 & \textit{Dispatch CSC} & 17 & 8 & 294 & 39 \\ \hline
	5 & NO  & 15693 & \textit{Dispatch CSC} & 22 & 1 & 106 & 6\\
	\hline
	6 & NO & 34464 & \textit{Flight Operations} & 55 & 10 & 96  & 6\\
	\hline
	7 & NO  & 15505 & \textit{Dispatch CSC} & 0 & 3 & 287 & 40\\
	\hline
	8 & YES & 13753 & \textit{Ground Operations} & -20 & 41 & 339 & 38\\
	\hline
	9 & NO & 5905 & \textit{Dispatch CSC}  & 51 & 24 & 103 & 8\\
	\hline
\end{tabularx}
\end{center}
\end{sidewaystable}

\begin{equation}\label{eqn:rp} 
    \begin{aligned}
       r = \{0,1,2,...\} = \{\,q \mid q \text{ is non-negative}\,\}
    \end{aligned}
\end{equation}
Fig.~\ref{fig:Hashgraph_rec} represents a snapshot in a dynamic and continuous timestamp horizon that infinitely increases the set $r$ as new disruptions occur for different functional roles in the AOCC. Thus, any number in the trace sequence $r$ that is not visible in Fig.~\ref{fig:Hashgraph_rec} is further up along the resolution horizon. Note that the circles shown in Fig.~\ref{fig:Hashgraph_rec} are unlabeled and all have the same size and color prior to achieving consensus on the Hashgraph platform. 

Table~\ref{tab:hashgraph_summaries} reveals the summaries of the first ten disruption resolution events (i.e. transactions) in the consensus recovery plan shown in Fig.~\ref{fig:Hashgraph_rec}. Of the ten disruption resolution events, four of them (i.e. $2,3,4,8$) are famous witnesses. Thus, the first (parent) disruption resolution events registered by intelligent domain managers on the Hashgraph ledger are for disrupted flight schedules with the following identification numbers (i.e. flight id) respectively: $2201, 11670, 12388,$ and $13753$. The highest tactical delay (i.e. delay applied before aircraft boarding and departure) of $61 mins$ was estimated for the recovery (or management) of disrupted flight $1536$ by the Fuel Management functional role. Conversely, the least tactical delay estimated among the ten disrupted flights from Table~\ref{tab:hashgraph_summaries} represents the execution of flight $13753$ twenty minutes earlier than originally planned by the Ground Operations functional role in the AOCC. Congruently, the highest strategic delay (i.e. delay applied after aircraft arrival and deplaning) of $40 mins$ was estimated for the recovery of disrupted flight $15505$ by the Dispatch CSC domain manager. Furthermore, the lowest strategic delay estimated among the ten disrupted flight schedules, highlighted in Table~\ref{tab:hashgraph_summaries}, affirms the arrival of flight $11670$ fourteen minutes earlier than the originally scheduled arrival time prior to disruption, as estimated by the Flight Operations functional domain manager. The total tactical and strategic delay estimated for recovering the ten flights are $269 mins$ and $190 mins$ respectively. As such, the recovery plan (from the decentralized AI platform) for the first ten flights will cost the airline an average of $\$359.55$ per disrupted passenger, assuming $\$47/hr$ \citep{A4A2019} as the average value of a passenger's time on each disrupted flight and the passenger flies on all ten disrupted flights.    

\subsection{Performance}
\begin{figure}[b!]
	\centering
	\includegraphics[width=0.99\textwidth]{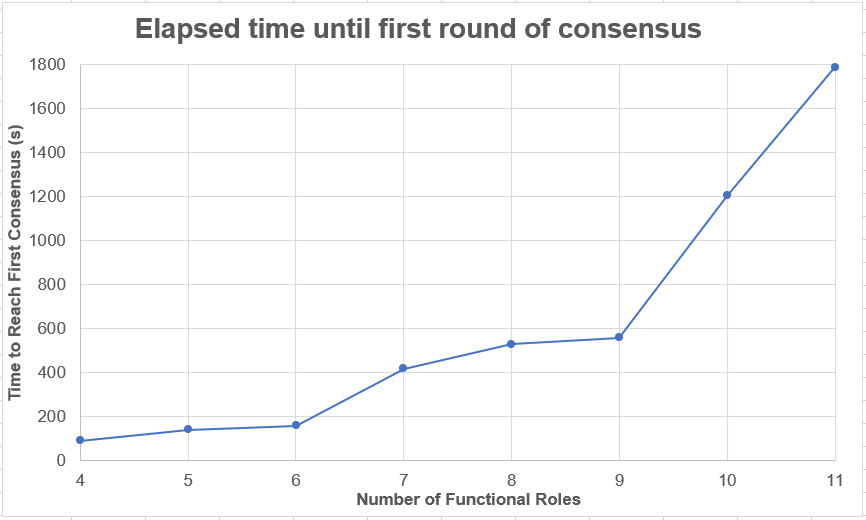}
	\caption{Elapsed time period before first round of consensus for an increasing number of functional roles in SWA-NOC}
	\label{fig:consensus_time}
\end{figure}

Fig.~\ref{fig:consensus_time} shows a plot of the period of time that elapsed before the first round of consensus is attained and registered on the Hashgraph platform, as the number of functional roles involved in disruption management (or operations recovery) is increased from four to eleven. Note that the number of functional roles, shown on the $x$-axis in Fig.~\ref{fig:consensus_time}, is increased based upon the alphabetical order of all the functional roles in SWA-NOC, thereby increasing the total number of disrupted flights for each scenario. As such, the scenario with four functional roles indicates an interaction of the following functional roles on the Hashgraph platform: Customer Hold, Dispatch CSC, Flight Operations, and Fuel Management, with a total of $1,057$ disrupted flights. In that vein, the scenario with five functional roles indicates an interaction of the Customer Hold, Dispatch CSC, Flight Operations, Fuel Management, and Ground Operations functional roles on the Hashgraph consensus platform, with a total of $2,741$ disrupted flights. Since the Hashgraph consensus timeline continues infinitely, we measure the time to reach first consensus in lieu. Hence, the time to reach first consensus, shown on the $y$-axis in Fig.~\ref{fig:consensus_time}, represents the duration from the start of the interaction amongst intelligent domain managers until the time when all famous witnesses in the first round are assigned a consensus position and corresponding timestamp, assuming the throughput that Hashgraph provides remains constant for all scenarios.  In essence, the time to reach first consensus represents the time it takes for the decentralized AI platform to come up with a credible recovery plan for a set of disrupted flight schedules from multiple functional roles. Fig.~\ref{fig:consensus_time} reveals that the rate at which the elapsed time to reach first consensus increases for interactions amongst nine to eleven functional roles is six times the rate at which the elapsed time to reach first consensus increases for interactions amongst four to six functional roles in the AOCC. As such, the minimum elapsed time to reach first consensus is $90 s$ during interaction for simultaneously-integrated recovery amongst four functional roles, and the maximum elapsed time to reach first consensus is $1780 s$ during interaction for simultaneously-integrated recovery amongst all eleven functional roles in SWA-NOC.

\section{Conclusion and Future Scope} \label{conclusion}
This paper provided an extensive review of fusion techniques for artificial intelligence (AI) and distributed ledger technology (DLT), and how they enabled the creation of a decentralized AI platform that simulates the integration and interaction of intelligent agents for simultaneously-integrated recovery during airline disruption management. Through a synthesis of AI models and a nonlinear parallel-chain DLT called Hashgraph, we developed and assessed the framework of an intelligent multi-agent system that can provide real-time schedule recovery plans based upon \textit{rules-of-thumb} provided by human experts in the airline operations control center (AOCC).

Furthermore, trust in human-AI collaboration for robust airline disruption management is contingent upon expedient verification and validation of the \textit{rules-of-thumb} that define input criteria for the intelligent multi-agent system. Thus, we achieved a semi-automatic platform that effectively eliminates the drawback of centralization for integration of functional roles (i.e. intelligent agents) in the AOCC, by engaging the decentralized and immutable nature of DLT to realize simultaneous interaction amongst intelligent agents during airline disruption management. We used the resultant decentralized AI platform to effect simultaneously-integrated recovery for an artificial network of disrupted real-world flight schedules across multiple functional roles in the AOCC, and obtained credible recovery plans within minutes of invoking the intelligent multi-agent system framework. In addition to executing in polynomial time, the decentralized AI platform proved to be effective when $100\%$ of the flights in the airline route network are disrupted, thereby outperforming existing platforms for airline disruption management that are only effective when no more than $65\%$ of flights are disrupted. To this effect, the findings from our research reveal the efficacy of simultaneous decision-making among intelligent functional roles in the AOCC through latest advancements in consensus mechanism design. 

While we present a novel data-driven design paradigm for enabling simultaneously-integrated recovery for airline disruption management, the data used for this research was provided by one U.S. airline that operates a point-to-point route network structure. As such, the AI models created from using this data may not completely represent industry-wide disruption management initiatives. Furthermore, the decentralized AI platform demonstrated in this paper uses a private DLT to describe interaction amongst functional roles in the AOCC at a U.S. airline. Thus, a future research direction is to use additional data from other air transportation stakeholders (e.g. airports) and airlines that operate a hub and spoke route network to create a consortium decentralized AI platform. This platform will bolster the efficacy of industry-wide disruption management initiatives such as FAA collaborative decision making and emerging operations such as UAS operations in the national airspace.       

\section*{Acknowledgement}
The authors would like to thank Blair Reeves, Chien Yu Chen, Kevin Wiecek, Jeff Agold, Dave Harrington, Rick Dalton, and Phil Beck, at Southwest Airlines Network Operations Control (SWA-NOC), for their expert inputs in abstracting the data used for this work.

\section*{Conflict of Interest}
All authors have no conflict of interest to report. 

\bibliography{mendeley}

\begin{thebibliography}{89}
\expandafter\ifx\csname natexlab\endcsname\relax\def\natexlab#1{#1}\fi
\providecommand{\url}[1]{\texttt{#1}}
\providecommand{\href}[2]{#2}
\providecommand{\path}[1]{#1}
\providecommand{\DOIprefix}{doi:}
\providecommand{\ArXivprefix}{arXiv:}
\providecommand{\URLprefix}{URL: }
\providecommand{\Pubmedprefix}{pmid:}
\providecommand{\doi}[1]{\href{http://dx.doi.org/#1}{\path{#1}}}
\providecommand{\Pubmed}[1]{\href{pmid:#1}{\path{#1}}}
\providecommand{\bibinfo}[2]{#2}
\ifx\xfnm\relax \def\xfnm[#1]{\unskip,\space#1}\fi
\bibitem[{{Airlines for America}(2019)}]{A4A2019}
\bibinfo{author}{{Airlines for America}} (\bibinfo{year}{2019}).
\newblock \bibinfo{title}{{U.S. Passenger Carrier Delay Costs}}.
\newblock \URLprefix
  \url{https://www.airlines.org/dataset/per-minute-cost-of-delays-to-u-s-airlines/#}.
\bibitem[{Allan \& Christopher(2005)}]{Allan2005}
\bibinfo{author}{Allan, M.}, \& \bibinfo{author}{Christopher, C.~K.}
  (\bibinfo{year}{2005}).
\newblock \bibinfo{title}{{Harmonising chorales by probabilistic inference}}.
\newblock In {\it \bibinfo{booktitle}{Advances in Neural Information Processing
  Systems}\/}.
\bibitem[{{Amadeus IT Group}(2016)}]{Amadeus2016}
\bibinfo{author}{{Amadeus IT Group}} (\bibinfo{year}{2016}).
\newblock \bibinfo{title}{{Airline Disruption Management}}.
\newblock \URLprefix
  \url{http://www.amadeus.com/documents/airline/airline-disruption-management/airline-disruption-management-whitepaper-2016.pdf}.
\bibitem[{Baird(2018)}]{Baird2018}
\bibinfo{author}{Baird, L.} (\bibinfo{year}{2018}).
\newblock \bibinfo{title}{{The swirlds hashgraph consensus algorithm: fair,
  fast, byzantine fault tolerance}}.
\newblock {\it \bibinfo{journal}{Swirlds Tech Report}\/},  {\it
  \bibinfo{volume}{01}\/}, \bibinfo{pages}{1--28}. \URLprefix
  \url{https://www.swirlds.com/downloads/SWIRLDS-TR-2016-01.pdf}.
  \DOIprefix\doi{10.1021/jp4051403}.
\bibitem[{Baird et~al.(2018)Baird, Harmon \& Madsen}]{Baird2018a}
\bibinfo{author}{Baird, L.}, \bibinfo{author}{Harmon, M.}, \&
  \bibinfo{author}{Madsen, P.} (\bibinfo{year}{2018}).
\newblock \bibinfo{title}{{Hedera: A governing council and public hashgraph
  network - The trust layer of the internet}}.
\newblock {\it \bibinfo{journal}{Whitepaper}\/},  (pp. \bibinfo{pages}{1--27}).
  \URLprefix
  \url{https://s3.amazonaws.com/hedera-hashgraph/hh-whitepaper-v1.1-180518.pdf%0Ahttps://www.hedera.com/hh-whitepaper-v2.0-300819.pdf}.
\bibitem[{Baird \& Luykx(2020)}]{Baird2020}
\bibinfo{author}{Baird, L.}, \& \bibinfo{author}{Luykx, A.}
  (\bibinfo{year}{2020}).
\newblock \bibinfo{title}{{The Hashgraph Protocol: Efficient Asynchronous BFT
  for High-Throughput Distributed Ledgers}}.
\newblock In {\it \bibinfo{booktitle}{2020 International Conference on
  Omni-layer Intelligent Systems (COINS)}\/} (pp. \bibinfo{pages}{1--7}).
\newblock \bibinfo{publisher}{IEEE}.
\newblock \URLprefix \url{https://ieeexplore.ieee.org/document/9191430/}.
  \DOIprefix\doi{10.1109/COINS49042.2020.9191430}.
\bibitem[{Ball et~al.(2010)Ball, Barnhart, Dresner, Hansen, Neels, Odoni,
  Peterson, Sherry, Trani \& Zou}]{Ball2010}
\bibinfo{author}{Ball, M.}, \bibinfo{author}{Barnhart, C.},
  \bibinfo{author}{Dresner, M.}, \bibinfo{author}{Hansen, M.},
  \bibinfo{author}{Neels, K.}, \bibinfo{author}{Odoni, A.},
  \bibinfo{author}{Peterson, E.}, \bibinfo{author}{Sherry, L.},
  \bibinfo{author}{Trani, A.~A.}, \& \bibinfo{author}{Zou, B.}
  (\bibinfo{year}{2010}).
\newblock \bibinfo{title}{{Total delay impact study: a comprehensive assessment
  of the costs and impacts of flight delay in the United States}}.
\newblock {\it \bibinfo{journal}{The national center of excellence NEXTOR}\/},
  (p. \bibinfo{pages}{2015}). \URLprefix
  \url{https://rosap.ntl.bts.gov/view/dot/6234}.
\bibitem[{Ball et~al.(2006)Ball, Barnhart, Nemhauser \& Odoni}]{Ball2006}
\bibinfo{author}{Ball, M.}, \bibinfo{author}{Barnhart, C.},
  \bibinfo{author}{Nemhauser, G.}, \& \bibinfo{author}{Odoni, A.}
  (\bibinfo{year}{2006}).
\newblock \bibinfo{title}{{Air Transportation : Irregular Operations and
  Control}}.
\newblock {\it \bibinfo{journal}{Handbooks of Operations Research and
  Management}\/},  (pp. \bibinfo{pages}{1--71}).
\bibitem[{Ball et~al.(2001)Ball, Chen, Hoffman \& Vossen}]{Ball2001a}
\bibinfo{author}{Ball, M.~O.}, \bibinfo{author}{Chen, C.-Y.},
  \bibinfo{author}{Hoffman, R.}, \& \bibinfo{author}{Vossen, T.}
  (\bibinfo{year}{2001}).
\newblock \bibinfo{title}{{Collaborative Decision Making in Air Traffic
  Management: Current and Future Research Directions}}.
\newblock \DOIprefix\doi{10.1007/978-3-662-04632-6{\_}2}.
\bibitem[{Barker \& Roginsky(2012)}]{Barker2012}
\bibinfo{author}{Barker, E.~B.}, \& \bibinfo{author}{Roginsky, A.}
  (\bibinfo{year}{2012}).
\newblock \bibinfo{title}{{Recommendation for Cryptographic Key Generation}}.
\newblock {\it \bibinfo{journal}{NIST Special Publication 800-133}\/}, .
  \DOIprefix\doi{10.6028/NIST.SP.800-133}.
\bibitem[{Barnhart(2009)}]{Barnhart2009}
\bibinfo{author}{Barnhart, C.} (\bibinfo{year}{2009}).
\newblock \bibinfo{title}{{Irregular Operations: Schedule Recovery and
  Robustness}}.
\newblock In {\it \bibinfo{booktitle}{The Global Airline Industry}\/} (pp.
  \bibinfo{pages}{253--274}).
\newblock \DOIprefix\doi{10.1002/9780470744734.ch9}.
\bibitem[{Barnhart et~al.(2003)Barnhart, Belobaba \& Odoni}]{Barnhart2003}
\bibinfo{author}{Barnhart, C.}, \bibinfo{author}{Belobaba, P.}, \&
  \bibinfo{author}{Odoni, A.} (\bibinfo{year}{2003}).
\newblock \bibinfo{title}{{Applications of Operations Research in the Air
  Transport Industry}}.
\newblock {\it \bibinfo{journal}{Transportation Science}\/},  {\it
  \bibinfo{volume}{37}\/}, \bibinfo{pages}{368--391}.
  \DOIprefix\doi{10.1287/trsc.37.4.368.23276}.
\bibitem[{Ben{\v{c}}i{\'{c}} \& {\v{Z}}arko(2018)}]{Bencic2018}
\bibinfo{author}{Ben{\v{c}}i{\'{c}}, F.~M.}, \& \bibinfo{author}{{\v{Z}}arko,
  I.~P.} (\bibinfo{year}{2018}).
\newblock \bibinfo{title}{{Distributed Ledger Technology: Blockchain Compared
  to Directed Acyclic Graph}}.
\newblock In {\it \bibinfo{booktitle}{Proceedings - International Conference on
  Distributed Computing Systems}\/}.
\newblock \DOIprefix\doi{10.1109/ICDCS.2018.00171}.
\bibitem[{Blakley(1979)}]{Blakley1979}
\bibinfo{author}{Blakley, G.~R.} (\bibinfo{year}{1979}).
\newblock \bibinfo{title}{{Safeguarding cryptographic keys}}.
\newblock In {\it \bibinfo{booktitle}{1979 International Workshop on Managing
  Requirements Knowledge, MARK 1979}\/}.
\newblock \DOIprefix\doi{10.1109/MARK.1979.8817296}.
\bibitem[{Bouarfa et~al.(2018)Bouarfa, M{\"{u}}ller \& Blom}]{Bouarfa2018}
\bibinfo{author}{Bouarfa, S.}, \bibinfo{author}{M{\"{u}}ller, J.}, \&
  \bibinfo{author}{Blom, H.} (\bibinfo{year}{2018}).
\newblock \bibinfo{title}{{Evaluation of a Multi-Agent System approach to
  airline disruption management}}.
\newblock {\it \bibinfo{journal}{Journal of Air Transport Management}\/}, .
  \DOIprefix\doi{10.1016/j.jairtraman.2018.05.009}.
\bibitem[{Bratu \& Barnhart(2005)}]{Bratu2005}
\bibinfo{author}{Bratu, S.}, \& \bibinfo{author}{Barnhart, C.}
  (\bibinfo{year}{2005}).
\newblock \bibinfo{title}{{An Analysis of Passenger Delays Using Flight
  Operations and Passenger Booking Data}}.
\newblock {\it \bibinfo{journal}{Air Traffic Control Quartely}\/},  {\it
  \bibinfo{volume}{13}\/}, \bibinfo{pages}{1--27}.
\bibitem[{Castro et~al.(2014)Castro, Paula, Eugenio \& Oliveira}]{Castro2014}
\bibinfo{author}{Castro, A. J.~M.}, \bibinfo{author}{Paula, A.},
  \bibinfo{author}{Eugenio, R.}, \& \bibinfo{author}{Oliveira, E.}
  (\bibinfo{year}{2014}).
\newblock {\it \bibinfo{title}{{Studies in Computational Intelligence 562 Ana
  Paula Rocha A New Approach for Disruption Management in Airline Operations
  Control}}\/}.
\bibitem[{Castro(2006)}]{Castro2006}
\bibinfo{author}{Castro, A. J. M.~D.} (\bibinfo{year}{2006}).
\newblock \bibinfo{title}{{Designing a Multi-Agent System for Monitoring and
  Operations Recovery for Airline Operations Control Centre}}.
\newblock {\it \bibinfo{journal}{The Lancet}\/},  {\it
  \bibinfo{volume}{367}\/}, \bibinfo{pages}{723}.
  \DOIprefix\doi{10.1016/S0140-6736(06)68290-1}.
\bibitem[{Castro \& Liskov(2002)}]{Castro2002}
\bibinfo{author}{Castro, M.}, \& \bibinfo{author}{Liskov, B.}
  (\bibinfo{year}{2002}).
\newblock \bibinfo{title}{{Practical Byzantine Fault Tolerance and Proactive
  Recovery}}.
\newblock {\it \bibinfo{journal}{ACM Transactions on Computer Systems}\/},
  {\it \bibinfo{volume}{20}\/}, \bibinfo{pages}{398--461}.
  \DOIprefix\doi{10.1145/571637.571640}.
\bibitem[{Choi et~al.(2018)Choi, Wallace \& Wang}]{Choi2018}
\bibinfo{author}{Choi, T.~M.}, \bibinfo{author}{Wallace, S.~W.}, \&
  \bibinfo{author}{Wang, Y.} (\bibinfo{year}{2018}).
\newblock \bibinfo{title}{{Big Data Analytics in Operations Management}}.
\newblock {\it \bibinfo{journal}{Production and Operations Management}\/},
  {\it \bibinfo{volume}{27}\/}, \bibinfo{pages}{1868--1883}.
  \DOIprefix\doi{10.1111/poms.12838}.
\bibitem[{Chui et~al.(2018)Chui, Manyika, Miremadi, Henke, Chung, Nel \&
  Malhotra}]{Chui2018}
\bibinfo{author}{Chui, M.}, \bibinfo{author}{Manyika, J.},
  \bibinfo{author}{Miremadi, M.}, \bibinfo{author}{Henke, N.},
  \bibinfo{author}{Chung, R.}, \bibinfo{author}{Nel, P.}, \&
  \bibinfo{author}{Malhotra, S.} (\bibinfo{year}{2018}).
\newblock \bibinfo{title}{{Notes from the AI frontier. Insights from hundreds
  of use cases}}, .
\newblock (p.~\bibinfo{pages}{36}). \URLprefix
  \url{https://www.mckinsey.com/mgi/}.
\bibitem[{Clarke(1998)}]{Clarke1998}
\bibinfo{author}{Clarke, M. D.~D.} (\bibinfo{year}{1998}).
\newblock \bibinfo{title}{{Irregular airline operations: a review of the
  state-of-the-practice in airline operations control centers}}.
\newblock {\it \bibinfo{journal}{Journal of Air Transport Management}\/}, .
  \DOIprefix\doi{10.1016/S0969-6997(98)00012-X}.
\bibitem[{Clausen et~al.(2010)Clausen, Larsen, Larsen \&
  Rezanova}]{Clausen2010}
\bibinfo{author}{Clausen, J.}, \bibinfo{author}{Larsen, A.},
  \bibinfo{author}{Larsen, J.}, \& \bibinfo{author}{Rezanova, N.~J.}
  (\bibinfo{year}{2010}).
\newblock \bibinfo{title}{{Disruption management in the airline
  industry-Concepts, models and methods}}.
\newblock {\it \bibinfo{journal}{Computers and Operations Research}\/},  {\it
  \bibinfo{volume}{37}\/}, \bibinfo{pages}{809--821}.
  \DOIprefix\doi{10.1016/j.cor.2009.03.027}.
\bibitem[{Correia et~al.(2011)Correia, Veronese, Neves \&
  Verissimo}]{Correia2011}
\bibinfo{author}{Correia, M.}, \bibinfo{author}{Veronese, G.~S.},
  \bibinfo{author}{Neves, N.~F.}, \& \bibinfo{author}{Verissimo, P.}
  (\bibinfo{year}{2011}).
\newblock \bibinfo{title}{{Byzantine consensus in asynchronous message-passing
  systems: A survey}}.
\newblock {\it \bibinfo{journal}{International Journal of Critical
  Computer-Based Systems}\/}, . \DOIprefix\doi{10.1504/IJCCBS.2011.041257}.
\bibitem[{Cover \& Thomas(2005)}]{Cover2005}
\bibinfo{author}{Cover, T.~M.}, \& \bibinfo{author}{Thomas, J.~A.}
  (\bibinfo{year}{2005}).
\newblock {\it \bibinfo{title}{{Elements of Information Theory}}\/}.
\newblock \DOIprefix\doi{10.1002/047174882X}.
\bibitem[{Crary(2021)}]{Crary2021}
\bibinfo{author}{Crary, K.} (\bibinfo{year}{2021}).
\newblock \bibinfo{title}{{Verifying the Hashgraph Consensus Algorithm}}, .
\newblock \URLprefix \url{http://arxiv.org/abs/2102.01167}.
\bibitem[{Dai et~al.(2019)Dai, Xu, Maharjan, Chen, He \& Zhang}]{Dai2019}
\bibinfo{author}{Dai, Y.}, \bibinfo{author}{Xu, D.}, \bibinfo{author}{Maharjan,
  S.}, \bibinfo{author}{Chen, Z.}, \bibinfo{author}{He, Q.}, \&
  \bibinfo{author}{Zhang, Y.} (\bibinfo{year}{2019}).
\newblock \bibinfo{title}{{Blockchain and Deep Reinforcement Learning Empowered
  Intelligent 5G beyond}}.
\newblock {\it \bibinfo{journal}{IEEE Network}\/}, .
  \DOIprefix\doi{10.1109/MNET.2019.1800376}.
\bibitem[{DeLaurentis(2005)}]{DeLaurentis2005a}
\bibinfo{author}{DeLaurentis, D.~A.} (\bibinfo{year}{2005}).
\newblock \bibinfo{title}{{Understanding transportation as system-of-systems
  design problem}}.
\newblock In {\it \bibinfo{booktitle}{43rd AIAA Aerospace Sciences Meeting and
  Exhibit - Meeting Papers}\/}.
\bibitem[{Dinh \& Thai(2018)}]{Dinh2018}
\bibinfo{author}{Dinh, T.~N.}, \& \bibinfo{author}{Thai, M.~T.}
  (\bibinfo{year}{2018}).
\newblock \bibinfo{title}{{AI and Blockchain: A Disruptive Integration}}.
\newblock {\it \bibinfo{journal}{Computer}\/}, .
  \DOIprefix\doi{10.1109/MC.2018.3620971}.
\bibitem[{Do \& Batzoglou(2008)}]{Do2008}
\bibinfo{author}{Do, C.~B.}, \& \bibinfo{author}{Batzoglou, S.}
  (\bibinfo{year}{2008}).
\newblock \bibinfo{title}{{What is the expectation maximization algorithm?}}
\newblock {\it \bibinfo{journal}{Nature Biotechnology}\/}, .
  \DOIprefix\doi{10.1038/nbt1406}.
\bibitem[{Fearing \& Barnhart(2011)}]{Fearing2011}
\bibinfo{author}{Fearing, D.}, \& \bibinfo{author}{Barnhart, C.}
  (\bibinfo{year}{2011}).
\newblock \bibinfo{title}{{Evaluating air traffic flow management in a
  collaborative decision-making environment}}.
\newblock {\it \bibinfo{journal}{Transportation Research Record}\/}, .
  \DOIprefix\doi{10.3141/2206-02}.
\bibitem[{Ghahramani(2015)}]{Ghahramani2015}
\bibinfo{author}{Ghahramani, Z.} (\bibinfo{year}{2015}).
\newblock \bibinfo{title}{{Probabilistic machine learning and artificial
  intelligence}}.
\newblock \DOIprefix\doi{10.1038/nature14541}.
\bibitem[{Hagel et~al.(2017)Hagel, Brown, De~Maar \& Wooll}]{Deloitte2017}
\bibinfo{author}{Hagel, J.}, \bibinfo{author}{Brown, J.~S.},
  \bibinfo{author}{De~Maar, A.}, \& \bibinfo{author}{Wooll, M.}
  (\bibinfo{year}{2017}).
\newblock \bibinfo{title}{{Southwest Airlines: Baker workgroup - Reducing
  disruption and delay to accelerate performance}}.
\bibitem[{Hao et~al.(2018)Hao, Yu, Zhiqiang, Zhen \& Dawu}]{Hao2018}
\bibinfo{author}{Hao, X.}, \bibinfo{author}{Yu, L.}, \bibinfo{author}{Zhiqiang,
  L.}, \bibinfo{author}{Zhen, L.}, \& \bibinfo{author}{Dawu, G.}
  (\bibinfo{year}{2018}).
\newblock \bibinfo{title}{{Dynamic practical byzantine fault tolerance}}.
\newblock In {\it \bibinfo{booktitle}{2018 IEEE Conference on Communications
  and Network Security, CNS 2018}\/}.
\newblock \DOIprefix\doi{10.1109/CNS.2018.8433150}.
\bibitem[{Heaton(2015)}]{Heaton2015}
\bibinfo{author}{Heaton, J.} (\bibinfo{year}{2015}).
\newblock {\it \bibinfo{title}{{Artificial Intelligence for Humans, Volume 3:
  Neural Networks and Deep Learning}}\/}.
\newblock \DOIprefix\doi{10.1007/978-1-4842-3450-1{\_}5}.
\bibitem[{Ibrahim(2017)}]{pyswirld2017}
\bibinfo{author}{Ibrahim, H.} (\bibinfo{year}{2017}).
\newblock \bibinfo{title}{{Py-swirld: Swirlds Consensus Algorithm in Python}}.
\newblock \URLprefix \url{https://github.com/Lapin0t/py-swirld#readme}.
\bibitem[{Janson et~al.(2008)Janson, Merkle \& Middendorf}]{Janson2008}
\bibinfo{author}{Janson, S.}, \bibinfo{author}{Merkle, D.}, \&
  \bibinfo{author}{Middendorf, M.} (\bibinfo{year}{2008}).
\newblock \bibinfo{title}{{A decentralization approach for swarm intelligence
  algorithms in networks applied to multi swarm PSO}}.
\newblock {\it \bibinfo{journal}{International Journal of Intelligent Computing
  and Cybernetics}\/}, . \DOIprefix\doi{10.1108/17563780810857112}.
\bibitem[{Jarrah et~al.(2000)Jarrah, Goodstein \& Narasimhan}]{Jarrah2000}
\bibinfo{author}{Jarrah, A.~I.}, \bibinfo{author}{Goodstein, J.}, \&
  \bibinfo{author}{Narasimhan, R.} (\bibinfo{year}{2000}).
\newblock \bibinfo{title}{{An Efficient Airline Re-Fleeting Model for the
  Incremental Modification of Planned Fleet Assignments}}.
\newblock {\it \bibinfo{journal}{Transportation Science}\/},  {\it
  \bibinfo{volume}{34}\/}, \bibinfo{pages}{349--363}.
  \DOIprefix\doi{10.1287/trsc.34.4.349.12324}.
\bibitem[{Jaynes(1957)}]{Jaynes1957}
\bibinfo{author}{Jaynes, E.~T.} (\bibinfo{year}{1957}).
\newblock \bibinfo{title}{{Information theory and statistical mechanics}}.
\newblock {\it \bibinfo{journal}{Physical Review}\/}, .
  \DOIprefix\doi{10.1103/PhysRev.106.620}.
\bibitem[{Jaynes(2003)}]{Jaynes2003}
\bibinfo{author}{Jaynes, E.~T.} (\bibinfo{year}{2003}).
\newblock \bibinfo{title}{{Probability Theory: The Logic of Science.}}
\newblock {\it \bibinfo{journal}{The Mathematical Intelligencer}\/},  {\it
  \bibinfo{volume}{27}\/}, \bibinfo{pages}{83--83}.
  \DOIprefix\doi{10.1007/BF02985800}.
\bibitem[{Karrer \& Newman(2009)}]{Karrer2009}
\bibinfo{author}{Karrer, B.}, \& \bibinfo{author}{Newman, M.~E.}
  (\bibinfo{year}{2009}).
\newblock \bibinfo{title}{{Random graph models for directed acyclic networks}}.
\newblock {\it \bibinfo{journal}{Physical Review E - Statistical, Nonlinear,
  and Soft Matter Physics}\/}, . \DOIprefix\doi{10.1103/PhysRevE.80.046110}.
\bibitem[{Keating et~al.(2003)Keating, Rogers, Unal, Dryer, Sousa-Poza,
  Safford, Peterson \& Rabadi}]{Keating2003}
\bibinfo{author}{Keating, C.}, \bibinfo{author}{Rogers, R.},
  \bibinfo{author}{Unal, R.}, \bibinfo{author}{Dryer, D.},
  \bibinfo{author}{Sousa-Poza, A.}, \bibinfo{author}{Safford, R.},
  \bibinfo{author}{Peterson, W.}, \& \bibinfo{author}{Rabadi, G.}
  (\bibinfo{year}{2003}).
\newblock \bibinfo{title}{{System of systems engineering}}.
\newblock {\it \bibinfo{journal}{EMJ - Engineering Management Journal}\/}, .
  \DOIprefix\doi{10.1080/10429247.2003.11415214}.
\bibitem[{Kiela et~al.(2016)Kiela, Bulat, Vero \& Clark}]{Kiela2016}
\bibinfo{author}{Kiela, D.}, \bibinfo{author}{Bulat, L.},
  \bibinfo{author}{Vero, A.~L.}, \& \bibinfo{author}{Clark, S.}
  (\bibinfo{year}{2016}).
\newblock \bibinfo{title}{{Virtual Embodiment: A Scalable Long-Term Strategy
  for Artificial Intelligence Research}}, .
\newblock \URLprefix \url{http://arxiv.org/abs/1610.07432}.
\bibitem[{Lamport et~al.(1982)Lamport, Shostak \& Pease}]{Lamport1982}
\bibinfo{author}{Lamport, L.}, \bibinfo{author}{Shostak, R.}, \&
  \bibinfo{author}{Pease, M.} (\bibinfo{year}{1982}).
\newblock \bibinfo{title}{{The Byzantine Generals Problem}}.
\newblock {\it \bibinfo{journal}{ACM Transactions on Programming Languages and
  Systems}\/},  {\it \bibinfo{volume}{4}\/}, \bibinfo{pages}{382--401}.
  \URLprefix \url{http://portal.acm.org/citation.cfm?doid=357172.357176}.
  \DOIprefix\doi{10.1145/357172.357176}.
\bibitem[{Lan et~al.(2006)Lan, Clarke \& Barnhart}]{Lan2006}
\bibinfo{author}{Lan, S.}, \bibinfo{author}{Clarke, J.-P.}, \&
  \bibinfo{author}{Barnhart, C.} (\bibinfo{year}{2006}).
\newblock \bibinfo{title}{{Planning for Robust Airline Operations: Optimizing
  Aircraft Routings and Flight Departure Times to Minimize Passenger
  Disruptions}}.
\newblock {\it \bibinfo{journal}{Transportation Science}\/},  {\it
  \bibinfo{volume}{40}\/}, \bibinfo{pages}{15--28}.
  \DOIprefix\doi{10.1287/trsc.1050.0134}.
\bibitem[{Lau et~al.(2000)Lau, Rubin, Smith \& Trajkovi{\'{c}}}]{Lau2000}
\bibinfo{author}{Lau, F.}, \bibinfo{author}{Rubin, S.~H.},
  \bibinfo{author}{Smith, M.~H.}, \& \bibinfo{author}{Trajkovi{\'{c}}, L.}
  (\bibinfo{year}{2000}).
\newblock \bibinfo{title}{{Distributed denial of service attacks}}.
\newblock {\it \bibinfo{journal}{Proceedings of the IEEE International
  Conference on Systems, Man and Cybernetics}\/}, .
  \DOIprefix\doi{10.1109/ICSMC.2000.886455}.
\bibitem[{Lewandowsky et~al.(2000)Lewandowsky, Mundy \& Tan}]{Lewandowsky2000}
\bibinfo{author}{Lewandowsky, S.}, \bibinfo{author}{Mundy, M.}, \&
  \bibinfo{author}{Tan, G.~P.} (\bibinfo{year}{2000}).
\newblock \bibinfo{title}{{The dynamics of trust: Comparing humans to
  automation}}.
\newblock {\it \bibinfo{journal}{Journal of Experimental Psychology:
  Applied}\/}, . \DOIprefix\doi{10.1037/1076-898X.6.2.104}.
\bibitem[{Li et~al.(2018)Li, Lin, Zheng, Wang, Liu, Cao, Wang \&
  Huang}]{Li2018}
\bibinfo{author}{Li, L.}, \bibinfo{author}{Lin, Y.~L.}, \bibinfo{author}{Zheng,
  N.~N.}, \bibinfo{author}{Wang, F.~Y.}, \bibinfo{author}{Liu, Y.},
  \bibinfo{author}{Cao, D.}, \bibinfo{author}{Wang, K.}, \&
  \bibinfo{author}{Huang, W.~L.} (\bibinfo{year}{2018}).
\newblock \bibinfo{title}{{Artificial intelligence test: a case study of
  intelligent vehicles}}.
\newblock {\it \bibinfo{journal}{Artificial Intelligence Review}\/}, .
  \DOIprefix\doi{10.1007/s10462-018-9631-5}.
\bibitem[{Liu et~al.(2019)Liu, Liu, Hansen, Pozdnukhov \& Zhang}]{Liu2019}
\bibinfo{author}{Liu, Y.}, \bibinfo{author}{Liu, Y.}, \bibinfo{author}{Hansen,
  M.}, \bibinfo{author}{Pozdnukhov, A.}, \& \bibinfo{author}{Zhang, D.}
  (\bibinfo{year}{2019}).
\newblock \bibinfo{title}{{Using machine learning to analyze air traffic
  management actions: Ground delay program case study}}.
\newblock {\it \bibinfo{journal}{Transportation Research Part E: Logistics and
  Transportation Review}\/}, . \DOIprefix\doi{10.1016/j.tre.2019.09.012}.
\bibitem[{Lordan et~al.(2016)Lordan, Sallan, Escorihuela \&
  Gonzalez-Prieto}]{Lordan2016}
\bibinfo{author}{Lordan, O.}, \bibinfo{author}{Sallan, J.~M.},
  \bibinfo{author}{Escorihuela, N.}, \& \bibinfo{author}{Gonzalez-Prieto, D.}
  (\bibinfo{year}{2016}).
\newblock \bibinfo{title}{{Robustness of airline route networks}}.
\newblock {\it \bibinfo{journal}{Physica A: Statistical Mechanics and its
  Applications}\/}, . \DOIprefix\doi{10.1016/j.physa.2015.10.053}.
\bibitem[{Magazzeni et~al.(2017)Magazzeni, Mcburney \& Nash}]{Magazzeni2017}
\bibinfo{author}{Magazzeni, D.}, \bibinfo{author}{Mcburney, P.}, \&
  \bibinfo{author}{Nash, W.} (\bibinfo{year}{2017}).
\newblock \bibinfo{title}{{Validation and verification of smart contracts: A
  research agenda}}.
\newblock {\it \bibinfo{journal}{Computer}\/}, .
  \DOIprefix\doi{10.1109/MC.2017.3571045}.
\bibitem[{Maher(2015)}]{Maher2015}
\bibinfo{author}{Maher, S.~J.} (\bibinfo{year}{2015}).
\newblock \bibinfo{title}{{A novel passenger recovery approach for the
  integrated airline recovery problem}}.
\newblock {\it \bibinfo{journal}{Computers and Operations Research}\/},  {\it
  \bibinfo{volume}{57}\/}, \bibinfo{pages}{123--137}.
  \DOIprefix\doi{10.1016/j.cor.2014.11.005}.
\bibitem[{Maier(1998)}]{Maier1998}
\bibinfo{author}{Maier, M.~W.} (\bibinfo{year}{1998}).
\newblock \bibinfo{title}{{Architecting Principles for Systems-of-Systems}}.
\newblock {\it \bibinfo{journal}{Systems Engineering}\/},  {\it
  \bibinfo{volume}{1}\/}, \bibinfo{pages}{267--284}.
  \DOIprefix\doi{10.1002/(SICI)1520-6858(1998)1:4<267::AID-SYS3>3.0.CO;2-D}.
\bibitem[{Maull et~al.(2017)Maull, Godsiff, Mulligan, Brown \&
  Kewell}]{maull2017}
\bibinfo{author}{Maull, R.}, \bibinfo{author}{Godsiff, P.},
  \bibinfo{author}{Mulligan, C.}, \bibinfo{author}{Brown, A.}, \&
  \bibinfo{author}{Kewell, B.} (\bibinfo{year}{2017}).
\newblock \bibinfo{title}{{Distributed ledger technology: Applications and
  implications}}.
\newblock {\it \bibinfo{journal}{Strategic Change}\/},  {\it
  \bibinfo{volume}{26}\/}, \bibinfo{pages}{481--489}.
  \DOIprefix\doi{10.1002/jsc.2148}.
\bibitem[{Maxmen(2018)}]{Maxmen2018}
\bibinfo{author}{Maxmen, A.} (\bibinfo{year}{2018}).
\newblock \bibinfo{title}{{AI researchers embrace Bitcoin technology to share
  medical data}}.
\newblock \DOIprefix\doi{10.1038/d41586-018-02641-7}.
\bibitem[{Medard \& Sawhney(2007)}]{Medard2007}
\bibinfo{author}{Medard, C.~P.}, \& \bibinfo{author}{Sawhney, N.}
  (\bibinfo{year}{2007}).
\newblock \bibinfo{title}{{Airline crew scheduling from planning to
  operations}}.
\newblock {\it \bibinfo{journal}{European Journal of Operational Research}\/},
  {\it \bibinfo{volume}{183}\/}, \bibinfo{pages}{1013--1027}.
  \DOIprefix\doi{10.1016/j.ejor.2005.12.046}.
\bibitem[{Moon(1996)}]{Moon1996}
\bibinfo{author}{Moon, T.~K.} (\bibinfo{year}{1996}).
\newblock \bibinfo{title}{{The expectation-maximization algorithm}}.
\newblock {\it \bibinfo{journal}{IEEE Signal Processing Magazine}\/}, .
  \DOIprefix\doi{10.1109/79.543975}.
\bibitem[{Mosier \& Skitka(2018)}]{Mosier2018}
\bibinfo{author}{Mosier, K.~L.}, \& \bibinfo{author}{Skitka, L.~J.}
  (\bibinfo{year}{2018}).
\newblock \bibinfo{title}{{Human decision makers and automated decision aids:
  Made for each other?}}
\newblock In {\it \bibinfo{booktitle}{Automation and Human Performance: Theory
  and Applications}\/}.
\newblock \DOIprefix\doi{10.1201/9781315137957}.
\bibitem[{Munro et~al.(2011)Munro, Toivonen, Webb, Buntine, Orbanz, Teh,
  Poupart, Sammut, Sammut, Blockeel, Rajnarayan, Wolpert, Gerstner, Page,
  Natarajan \& Hinton}]{Munro2011}
\bibinfo{author}{Munro, P.}, \bibinfo{author}{Toivonen, H.},
  \bibinfo{author}{Webb, G.~I.}, \bibinfo{author}{Buntine, W.},
  \bibinfo{author}{Orbanz, P.}, \bibinfo{author}{Teh, Y.~W.},
  \bibinfo{author}{Poupart, P.}, \bibinfo{author}{Sammut, C.},
  \bibinfo{author}{Sammut, C.}, \bibinfo{author}{Blockeel, H.},
  \bibinfo{author}{Rajnarayan, D.}, \bibinfo{author}{Wolpert, D.},
  \bibinfo{author}{Gerstner, W.}, \bibinfo{author}{Page, C.~D.},
  \bibinfo{author}{Natarajan, S.}, \& \bibinfo{author}{Hinton, G.}
  (\bibinfo{year}{2011}).
\newblock \bibinfo{title}{{Baum-Welch Algorithm}}.
\newblock In {\it \bibinfo{booktitle}{Encyclopedia of Machine Learning}\/}.
\newblock \DOIprefix\doi{10.1007/978-0-387-30164-8{\_}59}.
\bibitem[{Murakawa et~al.(2003)Murakawa, Adachi, Niino, Kasai, Takahashi,
  Takasuka \& Higuchi}]{Murakawa2003}
\bibinfo{author}{Murakawa, M.}, \bibinfo{author}{Adachi, T.},
  \bibinfo{author}{Niino, Y.}, \bibinfo{author}{Kasai, Y.},
  \bibinfo{author}{Takahashi, E.}, \bibinfo{author}{Takasuka, K.}, \&
  \bibinfo{author}{Higuchi, T.} (\bibinfo{year}{2003}).
\newblock \bibinfo{title}{{An AI-calibrated IF filter: A yield enhancement
  method with area and power dissipation reductions}}.
\newblock {\it \bibinfo{journal}{IEEE Journal of Solid-State Circuits}\/}, .
  \DOIprefix\doi{10.1109/JSSC.2002.808303}.
\bibitem[{Nakamoto(2008)}]{Nakamoto2008}
\bibinfo{author}{Nakamoto, S.} (\bibinfo{year}{2008}).
\newblock {\it \bibinfo{title}{{Bitcoin: A Peer-to-Peer Electronic Cash System
  | Satoshi Nakamoto Institute}}\/}.
\newblock \bibinfo{type}{Technical Report}.
\bibitem[{Ngo \& See(2012)}]{Ngo2012}
\bibinfo{author}{Ngo, T.~A.}, \& \bibinfo{author}{See, L.}
  (\bibinfo{year}{2012}).
\newblock \bibinfo{title}{{Calibration and validation of agent-based models of
  land cover change}}.
\newblock In {\it \bibinfo{booktitle}{Agent-Based Models of Geographical
  Systems}\/}.
\newblock \DOIprefix\doi{10.1007/978-90-481-8927-4{\_}10}.
\bibitem[{Nguyen et~al.(2018)Nguyen, Nguyen \& Nahavandi}]{Nguyen2018}
\bibinfo{author}{Nguyen, T.~T.}, \bibinfo{author}{Nguyen, N.~D.}, \&
  \bibinfo{author}{Nahavandi, S.} (\bibinfo{year}{2018}).
\newblock \bibinfo{title}{{Deep Reinforcement Learning for Multi-Agent Systems:
  A Review of Challenges, Solutions and Applications}}, .
\newblock (pp. \bibinfo{pages}{1--27}). \URLprefix
  \url{http://arxiv.org/abs/1812.11794}.
\bibitem[{Ogunsina et~al.(2021{\natexlab{a}})Ogunsina, Bilionis \&
  DeLaurentis}]{Ogunsina2021}
\bibinfo{author}{Ogunsina, K.}, \bibinfo{author}{Bilionis, I.}, \&
  \bibinfo{author}{DeLaurentis, D.} (\bibinfo{year}{2021}{\natexlab{a}}).
\newblock \bibinfo{title}{{Exploratory data analysis for airline disruption
  management}}.
\newblock {\it \bibinfo{journal}{Machine Learning with Applications}\/},  {\it
  \bibinfo{volume}{6}\/}, \bibinfo{pages}{100102}. \URLprefix
  \url{https://linkinghub.elsevier.com/retrieve/pii/S2666827021000517}.
  \DOIprefix\doi{10.1016/j.mlwa.2021.100102}.
\bibitem[{Ogunsina \& Okolo(2021)}]{Ogunsina2021b}
\bibinfo{author}{Ogunsina, K.}, \& \bibinfo{author}{Okolo, W.~A.}
  (\bibinfo{year}{2021}).
\newblock \bibinfo{title}{{Artificial Neural Network Modeling for Airline
  Disruption Management}}, .
\newblock \URLprefix \url{http://arxiv.org/abs/2104.02032}.
\bibitem[{Ogunsina et~al.(2021{\natexlab{b}})Ogunsina, Papamichalis \&
  DeLaurentis}]{Ogunsina2021a}
\bibinfo{author}{Ogunsina, K.}, \bibinfo{author}{Papamichalis, M.}, \&
  \bibinfo{author}{DeLaurentis, D.} (\bibinfo{year}{2021}{\natexlab{b}}).
\newblock \bibinfo{title}{{Uncertainty Quantification and Propagation for
  Airline Disruption Management}}, .
\newblock \URLprefix \url{http://arxiv.org/abs/2102.05147}.
\bibitem[{Okamura \& Yamada(2020)}]{Okamura2020}
\bibinfo{author}{Okamura, K.}, \& \bibinfo{author}{Yamada, S.}
  (\bibinfo{year}{2020}).
\newblock \bibinfo{title}{{Adaptive trust calibration for human-AI
  collaboration}}.
\newblock {\it \bibinfo{journal}{PLoS ONE}\/}, .
  \DOIprefix\doi{10.1371/journal.pone.0229132}.
\bibitem[{Olfati-Saber et~al.(2007)Olfati-Saber, Fax \&
  Murray}]{Olfati-Saber2007}
\bibinfo{author}{Olfati-Saber, R.}, \bibinfo{author}{Fax, J.~A.}, \&
  \bibinfo{author}{Murray, R.~M.} (\bibinfo{year}{2007}).
\newblock \bibinfo{title}{{Consensus and cooperation in networked multi-agent
  systems}}.
\newblock {\it \bibinfo{journal}{Proceedings of the IEEE}\/},  {\it
  \bibinfo{volume}{95}\/}, \bibinfo{pages}{215--233}.
  \DOIprefix\doi{10.1109/JPROC.2006.887293}.
\bibitem[{Oztemel \& Gursev(2020)}]{Oztemel2020}
\bibinfo{author}{Oztemel, E.}, \& \bibinfo{author}{Gursev, S.}
  (\bibinfo{year}{2020}).
\newblock \bibinfo{title}{{Literature review of Industry 4.0 and related
  technologies}}.
\newblock \DOIprefix\doi{10.1007/s10845-018-1433-8}.
\bibitem[{Panait \& Luke(2005)}]{Panait2005}
\bibinfo{author}{Panait, L.}, \& \bibinfo{author}{Luke, S.}
  (\bibinfo{year}{2005}).
\newblock \bibinfo{title}{{Cooperative multi-agent learning: The state of the
  art}}.
\newblock {\it \bibinfo{journal}{Autonomous Agents and Multi-Agent Systems}\/},
  . \DOIprefix\doi{10.1007/s10458-005-2631-2}.
\bibitem[{Petersen et~al.(2012)Petersen, S{\"{o}}lveling, Clarke, Johnson \&
  Shebalov}]{Petersen2012}
\bibinfo{author}{Petersen, J.~D.}, \bibinfo{author}{S{\"{o}}lveling, G.},
  \bibinfo{author}{Clarke, J.-P.}, \bibinfo{author}{Johnson, E.~L.}, \&
  \bibinfo{author}{Shebalov, S.} (\bibinfo{year}{2012}).
\newblock \bibinfo{title}{{An Optimization Approach to Airline Integrated
  Recovery}}.
\newblock {\it \bibinfo{journal}{Transportation Science}\/},  {\it
  \bibinfo{volume}{46}\/}, \bibinfo{pages}{482--500}. \URLprefix
  \url{http://pubsonline.informs.org/doi/abs/10.1287/trsc.1120.0414}.
  \DOIprefix\doi{10.1287/trsc.1120.0414}.
\bibitem[{Piccarozzi et~al.(2018)Piccarozzi, Aquilani \&
  Gatti}]{Piccarozzi2018}
\bibinfo{author}{Piccarozzi, M.}, \bibinfo{author}{Aquilani, B.}, \&
  \bibinfo{author}{Gatti, C.} (\bibinfo{year}{2018}).
\newblock \bibinfo{title}{{Industry 4.0 in management studies: A systematic
  literature review}}.
\newblock {\it \bibinfo{journal}{Sustainability (Switzerland)}\/}, .
  \DOIprefix\doi{10.3390/su10103821}.
\bibitem[{Pomerol(1997)}]{Pomerol1997}
\bibinfo{author}{Pomerol, J.} (\bibinfo{year}{1997}).
\newblock \bibinfo{title}{{Artificial intelligence and human decision making}}.
\newblock {\it \bibinfo{journal}{European Journal of Operational Research}\/},
  {\it \bibinfo{volume}{2217}\/}, \bibinfo{pages}{1--28}. \URLprefix
  \url{http://www.sciencedirect.com/science/article/pii/S0377221796003785}.
  \DOIprefix\doi{16/S0377-2217(96)00378-5}.
\bibitem[{Poole \& Mackworth(2010)}]{Poole2010}
\bibinfo{author}{Poole, D.~L.}, \& \bibinfo{author}{Mackworth, A.~K.}
  (\bibinfo{year}{2010}).
\newblock {\it \bibinfo{title}{{Artificial intelligence: Foundations of
  computational agents}}\/}.
\newblock \DOIprefix\doi{10.1017/CBO9780511794797}.
\bibitem[{Popov(2018)}]{Popov2018}
\bibinfo{author}{Popov, S.} (\bibinfo{year}{2018}).
\newblock \bibinfo{title}{{IOTA Whitepaper v1.4.3}}.
\newblock {\it \bibinfo{journal}{New Yorker}\/}, .
\bibitem[{Rauchs et~al.(2018)Rauchs, Glidden, Gordon, Pieters, Recanatini,
  Rostand, Vagneur \& Zhang}]{Rauchs2018}
\bibinfo{author}{Rauchs, M.}, \bibinfo{author}{Glidden, A.},
  \bibinfo{author}{Gordon, B.}, \bibinfo{author}{Pieters, G.~C.},
  \bibinfo{author}{Recanatini, M.}, \bibinfo{author}{Rostand, F.},
  \bibinfo{author}{Vagneur, K.}, \& \bibinfo{author}{Zhang, B.~Z.}
  (\bibinfo{year}{2018}).
\newblock \bibinfo{title}{{Distributed Ledger Technology Systems: A Conceptual
  Framework}}.
\newblock {\it \bibinfo{journal}{SSRN Electronic Journal}\/}, .
  \DOIprefix\doi{10.2139/ssrn.3230013}.
\bibitem[{Sage \& Cuppan(2001)}]{Sage2001}
\bibinfo{author}{Sage, A.}, \& \bibinfo{author}{Cuppan, C.}
  (\bibinfo{year}{2001}).
\newblock \bibinfo{title}{{On the Systems Engineering and Management of Systems
  of Systems and Federations of Systems}}.
\newblock {\it \bibinfo{journal}{Information Knowledge Systems Management}\/},
  .
\bibitem[{Salah et~al.(2019)Salah, Rehman, Nizamuddin \& Al-Fuqaha}]{Salah2019}
\bibinfo{author}{Salah, K.}, \bibinfo{author}{Rehman, M. H.~U.},
  \bibinfo{author}{Nizamuddin, N.}, \& \bibinfo{author}{Al-Fuqaha, A.}
  (\bibinfo{year}{2019}).
\newblock \bibinfo{title}{{Blockchain for AI: Review and open research
  challenges}}.
\newblock {\it \bibinfo{journal}{IEEE Access}\/}, .
  \DOIprefix\doi{10.1109/ACCESS.2018.2890507}.
\bibitem[{Sherali et~al.(2006)Sherali, Bish \& Zhu}]{Sherali2006}
\bibinfo{author}{Sherali, H.~D.}, \bibinfo{author}{Bish, E.~K.}, \&
  \bibinfo{author}{Zhu, X.} (\bibinfo{year}{2006}).
\newblock \bibinfo{title}{{Airline fleet assignment concepts, models, and
  algorithms}}.
\newblock {\it \bibinfo{journal}{European Journal of Operational Research}\/},
  {\it \bibinfo{volume}{172}\/}, \bibinfo{pages}{1--30}.
  \DOIprefix\doi{10.1016/j.ejor.2005.01.056}.
\bibitem[{Sinha(2006)}]{Sinha2006}
\bibinfo{author}{Sinha, S.} (\bibinfo{year}{2006}).
\newblock \bibinfo{title}{{On counting position weight matrix matches in a
  sequence, with application to discriminative motif finding}}.
\newblock In {\it \bibinfo{booktitle}{Bioinformatics}\/}.
\newblock \DOIprefix\doi{10.1093/bioinformatics/btl227}.
\bibitem[{Sousa et~al.(2016)Sousa, Teixeira, Cardoso \& Oliveira}]{Sousa2016}
\bibinfo{author}{Sousa, H.}, \bibinfo{author}{Teixeira, R.},
  \bibinfo{author}{Cardoso, H.~L.}, \& \bibinfo{author}{Oliveira, E.}
  (\bibinfo{year}{2016}).
\newblock \bibinfo{title}{{Airline Disruption Management}}, .
\newblock (pp. \bibinfo{pages}{398--405}).
\bibitem[{Swan(2015)}]{Swan2015}
\bibinfo{author}{Swan, M.} (\bibinfo{year}{2015}).
\newblock \bibinfo{title}{{Blockchain Thinking : the Brain as a Decentralized
  Autonomous Corporation [Commentary]}}.
\newblock \DOIprefix\doi{10.1109/MTS.2015.2494358}.
\bibitem[{Vanderweele \& Robins(2007)}]{Vanderweele2007}
\bibinfo{author}{Vanderweele, T.~J.}, \& \bibinfo{author}{Robins, J.~M.}
  (\bibinfo{year}{2007}).
\newblock \bibinfo{title}{{Directed acyclic graphs, sufficient causes, and the
  properties of conditioning on a common effect}}.
\newblock {\it \bibinfo{journal}{American Journal of Epidemiology}\/}, .
  \DOIprefix\doi{10.1093/aje/kwm179}.
\bibitem[{Vidal et~al.(2005)Vidal, Thollard, de~la Higuera, Casacuberta \&
  Carrasco}]{Vidal2005a}
\bibinfo{author}{Vidal, E.}, \bibinfo{author}{Thollard, F.},
  \bibinfo{author}{de~la Higuera, C.}, \bibinfo{author}{Casacuberta, F.}, \&
  \bibinfo{author}{Carrasco, R.~C.} (\bibinfo{year}{2005}).
\newblock \bibinfo{title}{{Probabilistic finite-state machines - Part I}}.
\newblock {\it \bibinfo{journal}{IEEE Transactions on Pattern Analysis and
  Machine Intelligence}\/}, . \DOIprefix\doi{10.1109/TPAMI.2005.147}.
\bibitem[{Watkins \& Dayan(1992)}]{Watkins1992}
\bibinfo{author}{Watkins, C. J. C.~H.}, \& \bibinfo{author}{Dayan, P.}
  (\bibinfo{year}{1992}).
\newblock \bibinfo{title}{{Q-learning}}.
\newblock {\it \bibinfo{journal}{Machine Learning}\/},  {\it
  \bibinfo{volume}{8}\/}, \bibinfo{pages}{279--292}. \URLprefix
  \url{http://link.springer.com/10.1007/BF00992698}.
  \DOIprefix\doi{10.1007/BF00992698}.
\bibitem[{Wright(2019)}]{Wright2019}
\bibinfo{author}{Wright, C.~S.} (\bibinfo{year}{2019}).
\newblock \bibinfo{title}{{Bitcoin: A Peer-to-Peer Electronic Cash System}}.
\newblock {\it \bibinfo{journal}{SSRN Electronic Journal}\/}, .
  \DOIprefix\doi{10.2139/ssrn.3440802}.
\bibitem[{Xia(2012)}]{Xia2012}
\bibinfo{author}{Xia, X.} (\bibinfo{year}{2012}).
\newblock \bibinfo{title}{{Position Weight Matrix, Gibbs Sampler, and the
  Associated Significance Tests in Motif Characterization and Prediction}}.
\newblock {\it \bibinfo{journal}{Scientifica}\/}, .
  \DOIprefix\doi{10.6064/2012/917540}.
\bibitem[{Ye et~al.(2020)Ye, Liu, Tian \& Wan}]{Ye2020}
\bibinfo{author}{Ye, B.}, \bibinfo{author}{Liu, B.}, \bibinfo{author}{Tian,
  Y.}, \& \bibinfo{author}{Wan, L.} (\bibinfo{year}{2020}).
\newblock \bibinfo{title}{{A methodology for predicting aggregate flight
  departure delays in airports based on supervised learning}}.
\newblock {\it \bibinfo{journal}{Sustainability (Switzerland)}\/}, .
  \DOIprefix\doi{10.3390/su12072749}.
\bibitem[{Zargar et~al.(2013)Zargar, Joshi \& Tipper}]{Zargar2013}
\bibinfo{author}{Zargar, S.~T.}, \bibinfo{author}{Joshi, J.}, \&
  \bibinfo{author}{Tipper, D.} (\bibinfo{year}{2013}).
\newblock \bibinfo{title}{{A survey of defense mechanisms against distributed
  denial of service (DDOS) flooding attacks}}.
\newblock {\it \bibinfo{journal}{IEEE Communications Surveys and Tutorials}\/},
  . \DOIprefix\doi{10.1109/SURV.2013.031413.00127}.

\end{thebibliography}

\newpage
\appendix
\section{Smart Contract Protocol for Consensus in Intelligent Multi-Agent System}\label{alg:consensus}
\begin{algorithm}
\caption{\textbf{\textit{Consensus Protocol}}}
\begin{algorithmic}[1]
\Require{$U_{M} = \{UTFM_{i}, ... , UTFM_{N}\}$} \Comment{AI models for stake estimation}
\Require$P_{M} = \{PTFM_{i}, ... , PTFM_{N}\}$ \Comment{AI models for impact estimation}
\Require{$H$} \Comment{Hashgraph DLT Algorithm}
\Procedure{IMASconsensus}{$X_{k}$}
\State $k \in M$ \Comment{where $M = \{1, ... , N\}$}
\For{$i = 1:N$}
\State $t_{i} \leftarrow UTFM_{i}(X_{k})$
\State $v_{i} \leftarrow \floor{-S_{i}\log_{2}{t_{i}}}$
\EndFor \Comment{where $t \in (0,1) \subset \mathbb{R}, \quad S \in \mathbb{Z}_{>0}$ }
\State $V = \{v_{1}, ... , v_{N}\}$
\State $p \leftarrow PTFM_{k}(X_{k})$, \quad  $stake \leftarrow V$ 
\State $r \leftarrow H(p, stake)$ \Comment{Consensus recovery plan}
\EndProcedure
\end{algorithmic}
\end{algorithm}

\end{document}